\documentclass{clv3}

\usepackage{hyperref}
\usepackage{xcolor}
\usepackage{graphicx}
\usepackage{url}
\usepackage{soul,color} 
\definecolor{darkblue}{rgb}{0, 0, 0.5}
\hypersetup{colorlinks=true,citecolor=darkblue, linkcolor=darkblue, urlcolor=darkblue}
\usepackage{siunitx}
\usepackage{cleveref}
\usepackage{framed}

\definecolor{shadecolor}{RGB}{180,180,180}

\runningtitle{Neural approaches to text readability}

\runningauthor{Martinc, Pollak, and Robnik-\v{S}ikonja}

\begin{document}

\title{Supervised and unsupervised neural approaches to text readability}

\author{Matej Martinc E-mail:matej.martinc@ijs.si.}
\affil{Jo\v{z}ef Stefan Institute, Ljubljana, Slovenia \\ Jo\v{z}ef Stefan International Postgraduate School, Ljubljana, Slovenia}

\author{Senja Pollak E-mail:senja.pollak@ijs.si.}
\affil{Jo\v{z}ef Stefan Institute, Ljubljana, Slovenia}

\author{Marko Robnik-\v{S}ikonja E-mail:marko.robnik@fri.uni-lj.si}
\affil{University of Ljubljana, Faculty of Computer and Information Science, Ljubljana, Slovenia}


\maketitle

\begin{snugshade*}
\noindent The  final reviewed  publication  was published in Computational Linguistics Journal, Volume 47,  Issue 1 - March 2021 and is available online at \url{https://doi.org/10.1162/coli_a_00398}
\end{snugshade*}

\begin{abstract}
We present a set of novel neural supervised and unsupervised approaches for determining the readability of documents. In the unsupervised setting, we leverage neural language models, whereas in the supervised setting, three different neural classification architectures are tested. We show that the proposed neural unsupervised approach is robust, transferable across languages and allows adaptation to a specific readability task and data set. By systematic comparison of several neural architectures on a number of benchmark and new labelled readability datasets in two languages, this study also offers a comprehensive analysis of different neural approaches to readability classification. We expose their strengths and weaknesses, compare their performance to current state-of-the-art classification approaches to readability, which in most cases still rely on extensive feature engineering, and propose possibilities for improvements.
\end{abstract}


\section{Introduction}
\label{sec-intro}

Readability is concerned with the relation between a given text and the cognitive load of a reader to comprehend it. This complex relation is influenced by many factors, such as a degree of lexical and syntactic sophistication, discourse cohesion, and background knowledge \citep{crossley2017predicting}. In order to simplify the problem of measuring readability, traditional readability formulas focused only on lexical and syntactic features expressed with statistical measurements, such as word length, sentence length, and word difficulty \citep{davison1982failure}. These approaches have been criticized because of their reductionism and weak statistical bases \citep{crossley2017predicting}. Another problem is their objectivity and cultural transferability since children from different environments master different concepts at different ages. For example, a word \textit{television} is quite long and contains many syllables but is well-known to most young children who live in families with a television.

With the development of novel natural language processing (NLP) techniques, several studies attempted to eliminate deficiencies of traditional readability formulas. These attempts include leveraging high-level textual features for readability modelling, such as semantic and discursive properties of texts. Among them, cohesion and coherence received the most attention, and several readability predictors based on these text features have been proposed (see Section \ref{related-work}). Nevertheless, none of them seems to predict the readability of the text as well as much simpler readability formulas mentioned above \cite{todirascu-etal-2016-cohesive}.

With the improvements in machine learning, the focus shifted once again, and most newer approaches consider readability as being a classification, regression, or a ranking task. Machine learning approaches build prediction models to predict human assigned readability scores based on several attributes and manually built features that cover as many text dimensions as possible \citep{schwarm2005reading, vajjala2012improving, petersen2009machine}. They generally yield better results than the traditional readability formulas and text cohesion based methods but require additional external resources, such as labelled readability datasets, which are scarce. Another problem is the transferability of these approaches between  different corpora and languages since the resulting feature sets do not generalize
well to different types of texts \cite{filighera2019automatic, xia2016text}.

Recently, deep neural networks \citep{Goodfellow2016} have shown impressive performance on many language-related tasks. In fact, they have achieved state-of-the-art performance in all semantic tasks where sufficient amounts of data were available \citep{Collobert2011,Zhang2015}. Even though very recently some neural approaches towards readability prediction have been proposed \citep{filighera2019automatic, nadeem2018estimating}, these type of studies are still relatively scarce, and further research is required in order to establish what type of neural architectures are the most appropriate for distinct readability tasks and datasets. Furthermore, language model features designed to measure lexical and semantic properties of text, which can be found in many of the readability studies \citep{schwarm2005reading, petersen2009machine, xia2016text}, are generated with traditional n-gram language models, even though language modelling has been drastically improved with the introduction of neural language models \citep{mikolov2011empirical}. 

The aim of the present study is two-fold. First, we propose a novel approach to readability measurement that takes into account neural language model statistics. This approach is unsupervised and requires no labelled training set but only a collection of texts from the given domain. We demonstrate that the proposed approach is capable of contextualizing the readability because of the trainable nature of neural networks and that it is transferable across different languages. In this scope, we propose a new measure of readability, RSRS (ranked sentence readability score), with good correlation with true readability scores.

Second, we experiment how different neural architectures with automatized feature generation can be used for readability classification and compare their performance to state-of-the-art classification approaches. Three distinct branches of neural architectures -- recurrent neural networks (RNN), hierarchical attention networks (HAN), and transfer learning techniques -- are tested on four gold standard readability corpora with good results. 

The paper is structured as follows. Section~\ref{related-work} addresses the related work on readability prediction. Section~\ref{sec:datasets} offers a thorough analysis of datasets used in our experiments, while in Section \ref{sec:unsupervised}, we present the methodology and results for the proposed unsupervised approach to readability prediction. The methodology and experimental results for the supervised approach are presented in Section \ref{sec:supervised}. We present conclusions and directions for further work in Section~\ref{sec:conclusion}.

\section{Related work}
\label{related-work}
Approaches to the automated measuring of readability try to find and assess factors that correlate well with human perception of readability. Several indicators, which measure different aspects of readability, have been proposed in the past and are presented in Section \ref{sec:readability-features}. These measures are used as features in newer approaches, which train machine learning models on texts with human-annotated readability levels so that they can predict readability levels on new unlabeled texts. Approaches, which rely on an extensive set of manually engineered features, are described in Section \ref{sec:traditional-classification}. Finally, Section \ref{sec:neural-classification} covers the approaches that tackle readability prediction with neural classifiers. Besides tackling the readability as a classification problem, several other supervised statistical approaches for readability prediction have been proposed in the past. They include regression \citep{sheehan2010generating}, SVM ranking \citep{ma2012ranking}, and graph-based methods \citep{jiang2015domain}, among many others. We do not cover these methods in the related work since they are not directly related to the proposed approach.

\subsection{Readability features}
\label{sec:readability-features}

Classical readability indicators can be roughly divided into five distinct groups: traditional, discourse cohesion, lexico-semantic, syntactic, and language model features. We describe them below.

\subsubsection{Traditional features}
\label{sec:formulas}
Traditionally, readability in texts was measured by statistical readability formulas, which try to construct a simple human-comprehensible formula with a good correlation to what humans perceive as the degree of readability. The simplest of them is average sentence length (ASL), though they take into account various other statistical factors, such as word length, and word difficulty. Most of these formulas were originally developed for English language but are also applicable to other languages with some modifications \citep{vskvorcevaluation}.

The Gunning fog index \citep{gunning1952technique} (GFI) estimates the years of formal education a person needs to understand the text on the first reading. It is calculated with the following expression:

\[ \textrm{GFI} = 0.4(\frac{totalWords} {totalSentences} + 100\frac{longWords} {totalSentences}), \] where \textit{longWords} are words longer than 7 characters. Higher values of the index indicate lower readability. 

Flesch reading ease \citep{kincaid1975derivation} (FRE) assigns higher values to more readable texts. It is calculated in the following way:

\[ \textrm{FRE} = 206.835 - 1.015 (\frac{totalWords} {totalSentences}) - 84.6 (\frac{totalSyllables} {totalWords}) \]

The values returned by the Flesch-Kincaid grade level \citep{kincaid1975derivation} (FKGL) correspond to the number of years of education generally required to understand the text for which the formula was calculated. The formula is defined as follows:

\[ \textrm{FKGL} = 0.39 (\frac{totalWords} {totalSentences}) + 11.8 (\frac{totalSyllables} {totalWords}) - 15.59 \]

Another readability formula that returns values corresponding to the years of education required to understand the text is Automated readability index \citep{smith1967automated} (ARI): 

\[\textrm{ARI}  = 4.71 (\frac{totalCharacters} {totalWords}) + 0.5 (\frac{totalWords} {totalSentences}) - 21.43 \]

Dale-Chall readability formula \citep{dale1948formula} (DCRF) requires a list of \num{3000} words that fourth-grade US students could reliably understand. Words that do not appear in this list are considered difficult. If the list of words is not available, it is possible to use the GFI approach and consider all the words longer than 7 characters as difficult. The following expression is used in calculation:  

\[\textrm{DCRF}  = 0.1579 (\frac{\textit{difficultWords}} {totalWords} * 100) + 0.0496 (\frac{totalWords} {totalSentences}) \]

The SMOG grade (Simple Measure of Gobbledygook) \cite{mc1969smog} is a readability formula originally used for checking health messages. Similar as FKGL and ARI, it roughly corresponds to the years of education needed to understand the text. It is calculated with the following expression: 

\[\textrm{SMOG}  = 1.0430 \sqrt{\textit{numberOfPolysyllables} \frac{30} {totalSentences}} + 3.1291, \] where the \textit{numberOfPolysyllables} is the number of words with three or more syllables.

We are aware of one study, which explored the transferability of these formulas across genres \citep{sheehan2013two}, and one study, which explored transferability across languages \citep{madrazo2020cross}. The study by \citet{sheehan2013two} concludes that mostly due to vocabulary specifics of different genres, traditional readability measures are not appropriate for cross-genre prediction, since they underestimate the complexity levels of literary texts and overestimate that of educational texts. The study by \citet{madrazo2020cross} on the other hand concludes that the readability level predictions for translations of the same text are rarely consistent when using these formulas. 

All of the above-mentioned readability measures were designed for the specific use on English texts. There are some rare attempts to adapt these formulas to other languages \citep{kandel1958application} or to create new formulas that could be used on languages other than English \citep{anderson1981analysing}. 

To show a multilingual potential of our approach, we address two languages in this study, English and Slovenian, a Slavic language with rich morphology and orders of magnitude fewer resources compared to English. 
For Slovenian, readability studies are scarce. \citet{vskvorcevaluation} researched how well the above statistical readability formulas work on Slovenian text by trying to categorize text from three distinct sources: children's magazines, newspapers and magazines for adults, and transcriptions of sessions of the National Assembly of Slovenia. Results of this study indicate that formulas which consider the length of words and/or sentences work better than formulas which rely on word lists. They also noticed that simple indicators of readability, such as percentage of adjectives and average sentence length, work quite well for Slovenian. To our knowledge, the only other study that employed readability formulas on Slovenian texts was done by \citet{vitez2014ugotavljanje}. Here the readability formulas were used as features in the author recognition task.

\subsubsection{Discourse cohesion features}
\label{sec:discourse-cohesion}

In the literature, we can find at least two distinct notions of discourse cohesion \citep{todirascu-etal-2016-cohesive}. First is the notion of \textbf{coherence}, defined as the ``semantic property of discourse, based on the interpretation of each sentence relative to the interpretation of other sentences'' \citep{van1977text}. Previous research which investigates this notion tries to determine whether a text can be interpreted as a coherent message and not just as a collection of unrelated sentences. This can be done by measuring certain observable features of the text, such as the repetition of content words or by analysis of words that explicitly express connectives (e.g., \textit{because, consequently, as a result}, etc.) \citep{sheehan2014textevaluator}. A somewhat more investigated, due to its easier operationalization, is the notion of \textbf{cohesion}, defined as ``a property of text represented by explicit formal grammatical ties (discourse connectives) and lexical ties that signal how utterances or larger text parts are related to each other''. 

According to \citet{todirascu-etal-2016-cohesive}, we can divide cohesion features into five distinct classes, outlined below: co-reference and anaphoric chain properties, entity density and entity cohesion features, lexical cohesion measures, and POS tag-based cohesion features.
\textbf{Co-reference and anaphoric chain properties} were first proposed by \citet{bormuth1969development}, who measured various characteristics of anaphora. These features include statistics, such as the average length of reference chains or the proportion of various types of mention (e.g., noun phrases, proper names, etc.) in the chain. \textbf{Entity density} features include statistics such as the total number of all/unique entities per document, the average number of all/unique entities per sentence, etc. These features were first proposed in \citet{feng2009cognitively} and \citet{feng2010comparison} who followed the theoretical line from \citet{halliday1976cohesion} and \citet{williams2006michael}. \textbf{Entity cohesion} features assess relative frequency of possible transitions between syntactic functions played by the same entity in adjacent sentences \citep{pitler2008revisiting}. \textbf{Lexical cohesion measures} include features such as the frequency of content word repetition across adjacent sentences \citep{sheehan2014textevaluator}, a Latent Semantic Analysis (LSA) based features for measuring the similarity of words and passages to each other proposed by \citet{landauer2011pearson}, or a measure called Lexical Tightness (LT) suggested by \citet{flor2013lexical}, defined as the mean value of the Positive Normalized Pointwise Mutual Information (PMI) for all pairs of content-word tokens in a text. The last category is \textbf{POS tag-based cohesion features} that measure the ratio of pronoun and article parts-of-speech, two crucial elements of cohesion \citep{todirascu-etal-2016-cohesive}.

\citet{todirascu-etal-2016-cohesive}, who analyzed 65 discourse features found in the readability literature, concluded, that they generally do not contribute much to the predictive power of text readability classifiers when compared to the traditional readability formulas or simple statistics such as sentence length. 

\subsubsection{Lexico-semantic features}

According to \citet{collins2014computational}, vocabulary knowledge is an important aspect of reading comprehension, and lexico-semantic features measure the difficulty of vocabulary in the text. A common feature is \textbf{Type-token ratio (TTR)}, which measures the ratio between the number of unique words and the total number of words in a text. The length of the text influences TTR; therefore, several corrections, which produce a more unbiased representation, such as Root TTR and Corrected TTR, are also used for readability prediction.

Other frequently used features in classification approaches to readability are \textbf{n-gram lexical features}, such as word and character n-grams \citep{vajjala2012improving, xia2016text}. While \textbf{part of speech (POS) based lexical features} measure lexical variation (i.e. TTR of lexical items such as nouns, adjectives, verbs, adverbs and prepositions) and density (e.g., the percentage of content words and function words), \textbf{word-list based features} use external psycholinguistic and Second  Language Acquisition (SLA) resources, which contain information about which words and phrases are acquired at the specific age or English learning class.

\subsubsection{Syntactic Features}

Syntactic features measure the grammatical complexity of the text and can be divided into several categories. \textbf{Parse tree features} include features such as an average parse tree height or an average number of noun- or verb-phrases per sentence. \textbf{Grammatical relations features} include measures of grammatical relations between constituents in a sentence, such as the longest/average distance in the grammatical relation sets generated by the parser. \textbf{Complexity of syntactic unit features} measure the length of a syntactic unit at the sentence, clause (any structure with a subject and a finite verb) and T-unit level (one main clause plus any subordinate clause). Finally, \textbf{coordination and subordination features} measure the amount of coordination and subordination in the sentence and include features such as a number of clauses per T-unit or number of coordinate phrases per clause, etc.

\subsubsection{Language model features}
\label{sec:LMs}
The standard task of language modeling can be formally defined as predicting a probability distribution of words from the fixed size vocabulary $V$, for word $w\textsubscript{t+1}$, given the historical sequence $w\textsubscript{1:t} = [w_1,...,w_t]$. 
To measure its performance, traditionally a metric called perplexity is used. A language model $m$ is evaluated according to how well it predicts a separate test sequence of words $w\textsubscript{1:N} = [w_1,...,w_N]$. For this case, the perplexity (PPL) of the language model $m()$ is defined as:

\begin{equation}
\textrm{PPL} = 2^{-\frac{1}{N} \sum_{i=1}^{N}\log_2{m(w_{i})}},
\label{eq:PPL}
\end{equation}
\noindent 
where $m(w_{i})$ is the probability assigned to word $w_i$ by the language model $m$, and $N$ is the length of the sequence. The lower the perplexity score, the better the language model predicts the words in a document, i.e. the more predictable and aligned with the training set the text is.

All past approaches for readability detection that employ language modeling, leverage older n-gram language models rather than the newer neural language models. \citet{schwarm2005reading} train one n-gram language model for each readability class \textit{c} in the training dataset. For each text document \textit{d}, they calculate the likelihood ratio according to the following formula:

\[ LR(d,c) = \frac{P(d|c)P(c)}{\sum_{\overline{c} \neq c}^{} P(d|\overline{c})P(\overline{c})}, \] 
where $P(d|c)$ denotes the probability returned by the language model trained on texts labeled with class \textit{c}, and $P(d|\overline{c})$ denotes probability of $d$ returned by the language model trained on the class $\overline{c}$. Uniform prior probabilities of classes are assumed. The likelihood ratios are used as features in the classification model along with perplexities achieved by all the models.

In \citet{petersen2009machine}, three statistical language models (unigram, bigram and trigram) are trained on four external data resources: Britannica (adult), Britannica Elementary, CNN (adult) and CNN abridged. The resulting twelve n-gram language models are used to calculate perplexities of each target document. It is assumed that low perplexity scores calculated by language models trained on the adult level texts and high perplexity scores of language models trained on the elementary/abridged levels would indicate a high reading level, and high perplexity scores of language models trained on the adult level texts and low perplexity scores of language models trained on the elementary/abridged levels would indicate a low reading level.

\citet{xia2016text} train 1- to 5-gram word-based language models on the British National Corpus, and 25 POS-based 1- to 5-gram models on the five classes of the WeeBit corpus. Language models' log-likelihood and perplexity scores are used as features for the classifier. 


\subsection{Classification approaches based on feature engineering}
\label{sec:traditional-classification}

The above approaches measure readability in an unsupervised way, using the described features. Alternatively, we can predict the level of readability in a supervised way. These approaches usually require extensive feature engineering and also leverage many of the features described above.

One of the first classification approaches to readability was proposed by \citet{schwarm2005reading}. It relies on a Support Vector Machine (SVM) classifier trained on a WeeklyReader corpus\footnote{http://www.weeklyreader.com}, containing articles grouped into four classes according to the age of the target audience. Traditional, syntactic, and language model features are used in the model. This approach was extended and improved upon in \citet{petersen2009machine}. 

Altogether 155 traditional, discourse cohesion, lexico-semantic and syntactic features were used in an approach proposed by \citet{vajjala2018onestopenglish}, tested on a recently published OneStopEnglish corpus. Sequential Minimal Optimization (SMO) classifier with the linear kernel achieved the classification accuracy of 78.13\% for three readability classes (elementary, intermediate, and advanced reading level). 

A successful classification approach to readability was proposed by \citet{vajjala2012improving}. Their multi-layer perceptron classifier is trained on the WeeBit corpus \citep{vajjala2012improving} (see Section \ref{sec:datasets} for more information on WeeBit and other mentioned corpora). The texts were classified into five classes according to the age group they are targeting. For classification, the authors use 46 manually crafted traditional, lexico-semantic and syntactic features. For the evaluation, they trained the classifier on a train set consisting of 500 documents from each class and tested it on a balanced test set of 625 documents (containing 125 documents per each class). They report 
93.3\% accuracy on the test set\footnote{A later research by \citet{xia2016text} called the validity of the published experimental results into question; therefore, the reported 93.3\% accuracy might not be the objective state-of-the-art result for readability classification.}. 

Another set of experiments on the WeeBit corpus was conducted by \citet{xia2016text} who conducted additional cleaning of the corpus since it contained some texts with broken sentences and additional meta-information about the source of the text, such as copyright declaration and links, strongly correlated with the target labels. They use similar lexical, syntactic, and traditional features as \citet{vajjala2012improving} but add language modeling (see Section \ref{sec:LMs} for details) and discourse cohesion based features. Their SVM classifier achieves 80.3\% accuracy using the 5-fold cross-validation. This is one of the studies where the transferability of the classification models is tested. Authors used an additional CEFR (Common European Framework of Reference for Languages) corpus. This small dataset of CEFR-graded texts is tailored for learners of English \citep{council2001common} and also contains 5 readability classes. The SVM classifier trained on the WeeBit corpus and tested on the CEFR corpus achieved the classification accuracy of 23.3\%, hardly beating the majority classifier baseline. This low result was attributed to the differences in readability classes in both corpora, since WeeBit classes are targeting children of different age groups, and CEFR corpus classes are targeting mostly adult foreigners with different levels of English comprehension. However, this result is a strong indication that transferability of readability classification models across different types of texts is questionable.

Two other studies that deal with the multi-genre prospects of readability prediction were conducted by \citet{sheehan2013two} and \citet{napolitano2015online}. Both studies describe the problem in the context of the TextEvaluator Tool \citep{sheehan2010generating}, an online system for text complexity analysis. The system supports multi-genre readability prediction with the help of a two-stage prediction workflow, in which first the genre of the text is determined (as being informational, literary or mixed) and after that its readability level is predicted with an appropriate genre-specific readability prediction model. Similarly to the study above, this work also indicates that using classification models for cross-genre prediction is not feasible. 

When it comes to multi- and cross-lingual classification,  \citet{madrazo2020cross} explore the possibility of a cross-lingual readability assessment and show that their methodology called CRAS (Cross-lingual Readability Assessment Strategy), which includes building a classifier that employs a set of traditional, lexico-semantic, syntactic and discourse cohesion based features works well in a multilingual setting. They also show that classification for some low resource languages can be improved by including documents from a different language into the train set for a specific language. 


\subsection{Neural classification approaches}
\label{sec:neural-classification}

Recently, several neural approaches for readability prediction have been proposed. \citet{nadeem2018estimating} tested two different architectures on the WeeBit corpus regression task, namely sequential Gated recurrent unit (GRU) \citep{cho2014learning} based RNN with the attention mechanism and hierarchical RNNs \citep{yang2016hierarchical} with two distinct attention types: a more classic attention mechanism proposed by \citet{bahdanau2014neural}, and multi-head attention proposed by \citet{vaswani2017attention}. The results of the study indicate that hierarchical RNNs generally perform better than sequential. \citet{nadeem2018estimating} also show that neural networks can be a good alternative to more traditional feature-based models for readability prediction on texts shorter than 100 words, but do not perform that competitively on longer texts.

Another version of a hierarchical RNN with the attention mechanism was proposed by \citet{azpiazu-pera-2019-multiattentive}. Their system, named Vec2Read, is a multi-attentive RNN capable of leveraging hierarchical text structures with the help of word and sentence level attention mechanisms and a custom-built aggregation mechanism. They employed the network in a multilingual setting (on corpora containing Basque, Catalan, Dutch, English, French, Italian, and Spanish texts). Their conclusion was, that while the number of instances used for training has a strong effect on the overall performance of the system, no language-specific patterns emerged that would indicate that prediction of readability in some languages is harder than in others.

An even more recent neural approach for readability classification on the cleaned WeeBit corpus \citep{xia2016text} was proposed by \citet{filighera2019automatic}, who tested a set of different embedding models, word2vec \citep{mikolov2013distributed}, the uncased Common Crawl GloVe \citep{pennington2014glove}, ELMo \citep{peters2018deep}, and BERT \citep{devlin2018bert}. The embeddings were fed to either a recurrent or a convolutional neural network. The BERT-based approach from their work is somewhat similar to the BERT-based supervised classification approach proposed in this work. However, one main distinction is that no fine-tuning is conducted on the BERT model in their experiments, i.e. the extraction of embeddings is conducted on the pretrained BERT language model. Their best ELMo-based model with a bidirectional LSTM achieved an accuracy of 79.2\% on the development set, slightly lower than the accuracy of 80.3\% achieved by \citet{xia2016text} in the 5-fold cross-validation scenario. However, they did manage to improve on the state-of-the-art by an ensemble of all their models, achieving the accuracy of 81.3\%, and the macro averaged $F_1$-score of 80.6\%.

A somewhat different neural approach to readability classification was proposed by \citet{mohammadi2019text}, who tackled the problem with deep reinforcement learning, or more specifically, with a deep convolutional recurrent double dueling Q network \citep{wang2016dueling} using a limited window of 5 adjacent words. GloVe embeddings and statistical language models were used to represent the input text in order to eliminate the need for sophisticated NLP features. The model was used in a multilingual setting (on English and Persian datasets) and achieved performance comparable to the state-of-the-art on all of the datasets, among them also on the Weebit corpus (accuracy of 91\%). 

Finally, a recent study by \citet{deutsch2020linguistic} used predictions of HAN and BERT models as additional features in their SVM model that also employed a set of syntactic and lexico-semantic features. While they did manage to improve the performance of their SVM classifiers with the additional neural features, they concluded that additional syntactic and lexico-semantic features did not generally improve the predictions of the neural models.

\section{Datasets}
\label{sec:datasets}

In this section, we first present the datasets used in the experiments (Section \ref{sec:dataset-presentation}) and then conduct their preliminary analysis (Section \ref{sec:dataset-analysis}) in order to assess the feasibility of the proposed experiments. Dataset statistics are presented in Table \ref{table:corpus-stats}.

\subsection{Dataset presentation}
\label{sec:dataset-presentation}
All experiments are conducted on four corpora labelled with readability scores:

\begin{itemize}
    \item { \bf The WeeBit corpus}: The articles from WeeklyReader\footnote{\url{http://www.weeklyreader.com}} and BBC-Bitesize\footnote{\url{http://www.bbc.co.uk/bitesize}} are classified into five classes according to the age group they are targeting. The classes correspond to age groups between 7-8, 8-9, 9-10, 10-14 and 14-16. Three classes targeting younger audiences consist of articles from WeeklyReader, an educational newspaper that covers a  wide range of non-fiction topics, from science to current affairs. Two classes targeting older audiences consist of material from the BBC-Bitesize website, containing educational material categorized into topics that roughly match school subjects in the UK. In the original corpus of \citet{vajjala2012improving}, the classes are balanced and the corpus contains altogether \num{3125} documents, \num{625} per class. In our experiments, we followed recommendations of \citet{xia2016text} to fix broken sentences and remove additional meta information, such as copyright declaration and links, strongly correlated with the target labels. We reextracted the corpus from the HTML files according to the procedure described in \citet{xia2016text} and discarded some documents due to the lack of content after the extraction and cleaning process. The final corpus used in our experiments contains altogether \num{3000} documents, \num{600} per class. 
    \item { \bf The OneStopEnglish corpus} \citep{vajjala2018onestopenglish} contains aligned texts of three distinct reading levels (beginner, intermediate, and advanced) that were written specifically for English as Second Language (ESL) learners. The corpus was compiled over the period 2013-2016 from the weekly news lessons section of the language learning resources \url{onestopenglish.com}. The section contains articles sourced from the Guardian newspaper, which were rewritten by English teachers to target three levels of adult ESL learners (elementary, intermediate, and advanced). Overall, the document aligned parallel corpus consists of 189 texts, each written in three versions (567 in total). The corpus is freely available\footnote{\url{https://zenodo.org/record/1219041}}.
    \item { \bf The Newsela corpus} \citep{xu2015problems}. We use the version of the corpus from 29 January 2016 consisting of altogether \num{10786} documents, out of which we only used \num{9565} English documents. The corpus contains \num{1911} original English news articles and up to four simplified versions for every original article, i.e., each original news article has been manually rewritten up to 4 times by editors at Newsela, a company that produces reading materials for pre-college classroom use, in order to target children at different grade levels and help teachers prepare curricula that match the English language skills required at each grade level. The dataset is a document aligned parallel corpus of original and simplified versions corresponding to altogether eleven different imbalanced grade levels (from 2nd to 12th grade). 
    
    \item { \bf Corpus of Slovenian school books (Slovenian SB):} In order to test the transferability of the proposed approaches to other languages, a corpus of Slovenian school books was compiled. The corpus contains \num{3639665} words in \num{125} school books for nine grades of primary schools and four grades of secondary school. It was created with several aims, like studying different quality aspects of school books, extraction of terminology, and linguistic analysis. The corpus contains school books for sixteen distinct subjects with very different topics ranging from literature, music and history to math, biology and chemistry, but not in equal proportions, with readers being the largest type of school books included.
    
    While some texts were extracted from the Gigafida reference corpus of written Slovene \citep{logar2012korpusi}, most of the texts were extracted from PDF files. After the extraction, we first conduct some light manual cleaning on the extracted texts (i.e., removal of indices, copyright statements, references, etc.). Next, in order to remove additional noise (e.g., tips, equations, etc.), we apply a filtering script that relies on manually written rules for sentence extraction (e.g., a text is a sentence if it starts with an uppercase and ends with an end of sentence punctuation) to obtain only passages containing sentences. Final extracted texts come without structural information (e.g., where does a specific chapter end or start, which sentences constitute a paragraph, where are questions, etc.), since labelling the document structure would require a large amount of manual effort; therefore we did not attempt it for this research. 
    
    For supervised classification experiments, we split the school books into chunks twenty-five sentences long, in order to build a train and test set with a sufficient number of documents\footnote{Note that this chunking procedure might break the text cohesion and that topical similarities between chunks from the same chapter (or paragraphs) might have a positive effect on the performance of the classification. However, since the corpus does not contain any high-level structural information (e.g., the information about paragraph or chapter structure of a specific school book), no other more refined chunking method is possible.}. The length of twenty-five sentences was chosen due to size limitations of the BERT classifier, which can be fed documents that contain up to 512 byte-pair tokens \citep{kudo2018sentencepiece}\footnote{Note that BERT tokenizer employs byte-pair tokenization \citep{kudo2018sentencepiece}, which in some cases generates tokens that correspond to sub-parts of words rather than entire words. In case of Slovenian SB, 512 byte-pair tokens correspond to 306 word tokens on average.}, which on average translates to slightly less than 25 sentences.
    
 \end{itemize}
 
 \begin{table*}[!ht]
\small
\centering
\caption{Readability classes, number of documents, tokens per specific readability class and average tokens per document in each readability corpus.}
\begin{tabular}{llll}
  Readability class & \#documents & \#tokens & \#tokens per doc.\\\hline
  \multicolumn{4}{c}{\textbf{Wikipedia}}\\\hline
  simple & 130,000 & 10,933,710 & 84.11\\
  balanced & 130,000 & 10,847,108 & 83.44\\
  normal & 130,000 & 10,719,878 & 82.46\\
  \multicolumn{4}{c}{\textbf{OneStopEnglish}}\\\hline
  beginner & 189 & 100,800 & 533.33\\
  intermediate & 189 & 127,934 & 676.90\\
  advanced & 189 & 155,253 & 820.49\\
  \textbf{All} & 567 & 383,987 & 677.23 \\
  \multicolumn{4}{c}{\textbf{WeeBit}}\\\hline
  age 7-8 & 600 & 77,613 & 129.35\\
  age 8-9 & 600 & 100,491 & 167.49\\
  age 9-10 & 600 & 159,719 & 266.20\\
  age 10-14 & 600 & 89,548 & 149.25\\
  age 14-16 & 600 & 152,402 & 254.00\\
  \textbf{All} & 3,000 & 579,773 & 193.26\\
  \multicolumn{4}{c}{\textbf{Newsela}}\\\hline
  2nd grade & 224 & 74,428 & 332.27\\
  3rd grade & 500 & 197,992 & 395.98\\
  4th grade & 1,569 & 923,828 & 588.80\\
  5th grade & 1,342 & 912,411 & 679.89\\
  6th grade & 1,058 & 802,057 & 758.09\\
  7th grade & 1,210 & 979,471 & 809.48\\
  8th grade & 1,037 & 890,358 & 858.59\\
  9th grade & 750 & 637,784 & 850.38\\
  10th grade & 20 & 19,012 & 950.60\\
  11th grade & 2 & 1,130 & 565.00\\
  12th & 1,853 & 1,833,781 & 989.63\\
  \textbf{All} & 9,565 & 7,272,252 & 760.30\\
  &&& \\
  \multicolumn{4}{c}{\textbf{KRES-balanced}}\\\hline
  balanced & / & 2,402,263 & / \\
  \multicolumn{4}{c}{\textbf{Slovenian SB}}\\\hline
  1st-ps & 69 & 12,921 & 187.26\\
  2nd-ps & 146 & 30,296 & 207.51\\
  3rd-ps & 268 & 62,241 & 232.24\\
  4th-ps & 1,007 & 265,242 & 263.40\\
  5th-ps & 1,186 & 330,039 & 278.28\\
  6th-ps & 959 & 279,461 & 291.41\\
  7th-ps & 1,470 & 462,551 & 314.66\\
  8th-ps & 1,844 & 540,944 & 293.35\\
  9th-ps & 2,154 & 688,149 & 319.47\\
  1st-hs & 1,663 & 578,694 & 347.98\\
  2nd-hs & 590 & 206,147 & 349.40\\
  3rd-hs & 529 & 165,845 & 313.51\\
  4th-hs & 45 & 14,313 & 318.07\\
  \textbf{All} & 11,930 & 3,636,843 & 304.85\\

\vspace*{0.5cm}
 
\end{tabular}
\label{table:corpus-stats}
\end{table*}

Language models are trained on large corpora of texts. For this purpose, we used the following corpora.

\begin{itemize}
    \item { \bf Corpus of English Wikipedia} and {\bf Corpus of Simple Wikipedia} articles. We created three corpora for the use in our unsupervised English experiments\footnote{English Wikipedia and Simple Wikipedia dumps from 26th of January 2018 were used for the corpus construction.}:
    
    \begin{itemize}
        \item \textbf{Wiki-normal} contains \num{130000} randomly selected articles from the Wikipedia dump, which comprise of \num{489976} sentences and \num{10719878} tokens.
        \item \textbf{Wiki-simple} contains \num{130000} randomly selected articles from the Simple Wikipedia dump, which comprise of \num{654593} sentences and \num{10933710} tokens.
        \item \textbf{Wiki-balanced} contains \num{65000} randomly selected articles from the Wikipedia dump (dated 26 January 2018) and \num{65000} randomly selected articles from the Simple Wikipedia dump. Altogether the corpus comprises of  \num{571964} sentences and \num{10847108} tokens. 
    \end{itemize}{}
    
    \item { \bf KRES-balanced:}  KRES corpus \citep{logar2012korpusi} is a 100 million word balanced reference corpus of Slovenian language. 35\% of its content are books, 40\% periodicals, and 20\% internet texts. From this corpus we took all the available documents from two children magazines (Ciciban and Cicido), all documents from four teenager magazines (Cool, Frka, PIL plus and Smrklja), and documents from three magazines targeting adult audiences (\v{Z}ivljenje in tehnika, Radar, City magazine). With these texts, we built a corpus with approximately 2.4 million words. The corpus is balanced in a sense that about one-third of the sentences come from documents targeting children, one third is targeting teenagers, and the last third is targeting adults.
\end{itemize}

\subsection{Dataset analysis}
\label{sec:dataset-analysis}

Overall, there are several differences between our datasets:

\begin{itemize}
    \item \textbf{Language:} As already mentioned before, we have three English (Newsela, OneStopEnglish and WeeBit) and one Slovenian (Slovenian SB) test dataset.
    \item \textbf{Parallel corpora vs unaligned corpora:} Newsela and OneStopEnglish datasets are parallel corpora, which means that articles from different readability classes are semantically similar to each other. On the other hand, WeeBit and Slovenian SB datasets contain completely different articles in each readability class. While this might not affect traditional readability measures, which do not take semantic information into account, it might prove substantial for the performance of classifiers and the proposed language model based readability measures.
    \item \textbf{Length of documents:}  Another difference between Newsela and OneStopEnglish datasets on one side, and WeeBit and Slovenian SB dataset on the other is the length of dataset documents. While Newsela and OneStopEnglish datasets contain longer documents, on average about 760 and 677 words long, documents in the WeeBit and Slovenian SB corpora are on average about 193 and 305 words long, respectively.
    \item \textbf{Genre:} OneStopEnglish and Newsela datasets contain news articles, WeeBit is made of educational articles, and the Slovenian SB dataset is composed of school books. For training of the English language models, we use Wikipedia and Simple Wikipedia, which contain encyclopedia articles, and for Slovene language model training, we use the KRES-balanced corpus, which contains magazine articles. 
    \item \textbf{Target audience:}  OneStopEnglish is the only test dataset that specifically targets adult ESL learners and not children, as do other test datasets. When it comes to datasets used for language model training, KRES-balanced corpus is made of articles which target both adults and children. The problem with Wikipedia and Simple Wikipedia is that no specific target audience is addressed since articles are written by volunteers. In fact, using Simple Wikipedia as a dataset for the training of simplification algorithms has been criticized in the past due to lack of specific simplification guidelines, which are based only on the declarative statement that Simple Wikipedia was created for ``children and adults who are learning the English language'' \citep{xu2015problems}. This lack of guidelines also contributes to the decrease in the quality of simplification according to \citet{xu2015problems}, who found that the corpus can be noisy and that half of its sentences are not actual simplifications but rather copied from the original Wikipedia.    
    
\end{itemize}

\begin{table*}[h!!tp]
\centering
\caption{Scores of traditional readability indicators from \Cref{sec:formulas} for specific classes in the readability datasets.}
\resizebox{0.83 \textwidth}{!}{
\begin{tabular}{lrrrrrrr}
 Class & GFI & FRE & FKGL & ARI & DCRF & SMOG & ASL\\ \hline
\multicolumn{8}{c}{\textbf{Wikipedia}}\\\hline
simple & 11.80 & 62.20 & 8.27 & 14.08 & 11.40 & 11.40 & 16.90 \\
balanced & 13.49 & 56.17 & 9.70 & 15.86 & 12.53 & 12.53 & 19.54\\
normal & 15.53 & 49.16 & 11.47 & 18.06 & 13.89 & 13.89 & 23.10 \\
\multicolumn{8}{c}{\textbf{WeeBit}}\\\hline
age 7-8 & 6.91 & 83.41 & 3.82 & 8.83 & 7.83 & 7.83 & 10.23\\
age 8-9 & 8.45 & 76.68 & 5.34 & 10.33 & 8.87 & 8.87 & 12.89\\
age 9-10 & 10.30 & 69.88 & 6.93 & 12.29 & 10.01 & 10.01 & 15.69\\
age 10-14 & 9.94 & 75.35 & 6.34 & 11.20 & 9.67 & 9.67 & 16.64\\
age 14-16 & 11.76 & 66.61 & 8.09 & 13.56 & 10.81 & 10.81 & 18.86\\
\multicolumn{8}{c}{\textbf{OneStopEnglish}}\\\hline
beginner & 11.79 & 66.69 & 8.48 & 13.93 & 11.05 & 11.05 & 20.74\\
intermediate & 13.83 & 59.68 & 10.19 & 15.98 & 12.30 & 12.30 & 23.98\\
advanced & 15.35 & 54.84 & 11.54 & 17.65 & 13.22 & 13.22 & 26.90\\
\multicolumn{8}{c}{\textbf{Newsela}}\\\hline
2nd grade & 6.11 & 85.69 & 3.27 & 8.09 & 7.26 & 7.26 & 9.26\\
3rd grade & 7.24 & 80.92 & 4.27 & 9.30 & 7.94 & 7.94 & 10.72\\
4th grade & 8.58 & 76.05 & 5.40 & 10.50 & 8.88 & 8.88 & 12.72\\
5th grade & 9.79 & 71.76 & 6.47 & 11.73 & 9.68 & 9.68 & 14.81\\
6th grade & 11.00 & 67.46 & 7.53 & 12.99 & 10.47 & 10.47 & 16.92\\
7th grade & 12.11 & 62.71 & 8.54 & 14.12 & 11.26 & 11.26 & 18.46\\
8th grade & 13.05 & 60.37 & 9.38 & 15.19 & 11.83 & 11.83 & 20.81\\
9th grade & 14.20 & 55.00 & 10.46 & 16.37 & 12.70 & 12.70 & 22.17\\
10th grade & 14.15 & 55.70 & 10.60 & 16.50 & 12.83 & 12.83 & 23.33\\
11th grade & 15.70 & 56.41 & 11.05 & 16.96 & 12.77 & 12.77 & 24.75\\
12th grade & 14.52 & 55.58 & 10.71 & 16.70 & 12.79 & 12.79 & 23.69\\
&&&&&&& \\
\multicolumn{8}{c}{\textbf{KRES-balanced}}\\\hline
balanced & 12.72 & 29.20 & 12.43 & 14.88 & 14.08 & 14.08 & 15.81\\
\multicolumn{8}{c}{\textbf{Slovenian SB}}\\\hline
1st-ps & 9.54 & 31.70 & 10.38 & 11.72 & 11.12 & 11.12 & 7.63\\
2nd-ps & 9.49 & 34.90 & 10.11 & 11.34 & 11.26 & 11.26 & 8.37\\
3rd-ps & 10.02 & 32.89 & 10.61 & 11.78 & 11.80 & 11.80 & 9.31\\
4th-ps & 10.96 & 30.29 & 11.18 & 12.84 & 12.39 & 12.39 & 10.40\\
5th-ps & 11.49 & 28.13 & 11.62 & 13.33 & 12.79 & 12.79 & 11.02\\
6th-ps & 13.20 & 20.10 & 12.84 & 14.57 & 13.61 & 13.61 & 11.45\\
7th-ps & 12.94 & 22.97 & 12.61 & 14.52 & 13.64 & 13.64 & 12.24\\
8th-ps & 13.48 & 18.12 & 13.09 & 14.78 & 13.71 & 13.71 & 11.32\\
9th-ps & 13.69 & 19.26 & 13.13 & 15.07 & 13.94 & 13.94 & 12.27\\
1st-hs & 15.12 & 12.66 & 14.33 & 16.22 & 14.96 & 14.96 & 13.62\\
2nd-hs & 15.13 & 15.13 & 13.90 & 15.83 & 14.67 & 14.67 & 13.49\\
3rd-hs & 14.76 & 13.09 & 14.00 & 15.62 & 14.44 & 14.44 & 12.57\\
4th-hs & 14.66 & 14.39 & 13.64 & 15.54 & 14.03 & 14.03 & 11.62\\
\hline 
\end{tabular}
}
\label{table:datasets-stats}
\end{table*}

This diversity of the datasets limits ambitions of the study to offer general conclusions true across genres, languages, or datasets. On the other hand, it offers an opportunity to determine how specifics of each dataset affect each of the proposed readability predictors and also to determine the overall robustness of the applied methods. 

While many aspects differ from one dataset to another, there are also some common characteristics across all the datasets, which allow using the same prediction methods on all of them. These are mostly connected to the common techniques used in the construction of the readability datasets, no matter the language, genre, or target audience of the specific dataset. The creation of parallel simplification corpora (i.e. Newsela, OneStopEnglish, and Simple Wikipedia) generally involves three techniques, splitting (breaking a long sentence into shorter ones), deletion (removing unimportant parts of a sentence), and paraphrasing (rewriting a text into a simpler version via reordering, substitution, and occasionally expansion) \citep{feng2008text}. Even though there might be some subtleties involved (since what constitutes simplification for one type of user may not be appropriate for another) how these techniques are applied is rather general. Also, while there is no simplification used in the non-parallel corpora (WeeBit, Slovenian SB), the contributing authors were nevertheless instructed to write the text for a specific target group and adapt the writing style accordingly. In most cases, this leads to the same result, e.g., shorter less complex sentences and simpler vocabulary used in texts intended for younger or less fluently speaking audiences. 

The claim of commonality between datasets can be backed up by the fact, that even traditional readability indicators correlate quite well to human assigned readability, no matter the specific genre, language, or purpose of each dataset. Results in Table \ref{table:datasets-stats} demonstrate this point by showcasing readability scores of traditional readability formulas from Section \ref{sec:formulas}. We can see that the general pattern of increased difficulty on all datasets and for all indicators --- larger readability scores (or in case of FRE, smaller) are assigned to those classes of the dataset that contain texts written for older children or more advanced ESL learners. This suggests that multi-dataset, multi-genre and even multi-lingual readability prediction is feasible on the set of chosen datasets, even if only the shallow traditional readability indicators are used. 

However, the results do indicate that cross-genre or even cross-dataset readability prediction might be problematic since the datasets do not cover the same readability range according to the shallow prediction formulas (and also ground truth readability labels). For example, documents in the WeeBit 14-16 age group have scores very similar to the Newsela 6th grade documents, which means that a classifier trained on the WeeBit corpus might have a hard time classifying documents belonging to higher Newsela grades since the readability of these documents is lower than for the most complex documents in the WeeBit corpus according to all of the shallow readability indicators. For this reason, we opted not to perform any supervised cross-dataset or cross-genre experiments. Nevertheless, the problem of cross-genre prediction is important in the context of the proposed unsupervised experiments, since the genre discrepancy between the datasets used for training the language models and the datasets on which the models are employed, might influence the performance of the proposed language model based measures. A more detailed discussion on this topic is presented in Section \ref{sec-methodology-unsupervised}.   

The analysis in \Cref{table:datasets-stats} also confirms the findings by \citet{madrazo2020cross}, who have shown that cross-lingual readability prediction with shallow readability indicators is problematic. For example, if we compare the Newsela corpus and Slovenian SB corpus, which both cover roughly the same age group, we can see that for some readability indicators (FRE, FKGL, DCRF, and ASL) the values are on entirely different scales.

\section{Unsupervised neural approach}
\label{sec:unsupervised}

In this section, we explore how neural language models can be used for determining the readability of the text in an unsupervised way. In Section \ref{sec:NLMs}, we present the neural architectures used in our experiments, in Section \ref{sec-methodology-unsupervised}, we describe the methodology of the proposed approach, and in Section \ref{sec-unsupervised-experiments}, we present the conducted experiments.

\subsection{Neural language model architectures}
\label{sec:NLMs}

\citet{mikolov2011empirical} have shown that neural language models outperform n-gram language models by a high margin on large and also relatively small (less than 1 million tokens) datasets. The achieved differences in perplexity (see Eq. (\ref{eq:PPL})) are attributed to a richer historical contextual information available to neural networks, which are not limited to a small contextual window (usually of up to five previous words) as is the case of n-gram language models. In Section \ref{sec:LMs}, we mentioned some approaches that use n-gram language models for readability prediction. However, we are unaware of any approach that would employ deep neural network language models for determining the readability of a text.

In this research, we employ three neural architectures for language modelling. First are recurrent neural networks (RNN), which are suitable for modelling sequential data. At each time step $t$, the input vector $x_t$, and hidden state vector $h\textsubscript{t-1}$ are feed into the network, producing the next hidden vector state $h_t$ with the following recursive equation:

\[
h_t = f(Wx_t + Uh\textsubscript{t-1} + b),
\] 
where $f$ is a non-linear activation function, $W$ and $U$ are matrices representing weights of the input layer and hidden layer,  and $b$ is the bias vector. Learning long-range input dependencies with plain RNNs is problematic due to vanishing gradients \citep{bengio1994learning}, therefore, in practice, modified recurrent networks, such as Long short-term memory networks (LSTM) are used. In our experiments, we use the LSTM-based language model proposed by \citet{kim2016character}. This architecture is adapted to language modelling of morphologically rich languages, such as Slovenian, by employing an additional character-level convolutional neural network (CNN). The convolutional level learns a character structure of words and is connected to the LSTM-based layer, which produces predictions at the word level.

 \citet{bai2018empirical} introduced a new sequence modelling architecture based on convolution, called temporal convolutional network (TCN), which is also employed in our experiments. TCN uses causal convolution operations, which make sure that there is no information leakage from future time steps to the past. This and the fact that TCN takes a sequence as an input and maps it into an output sequence of the same size makes this architecture appropriate for language modelling. TCNs are capable of leveraging long contexts by using a very deep network architecture and a hierarchy of dilated convolutions. A single dilated convolution operation $F$ on element $s$ of the 1-dimensional sequence $x$ can  be defined with the following equation:

\[F(s) = (x * \textsubscript{d}f)(s) = \sum_{i=0}^{k-1} f(i) \cdot x_{s-d\cdot i},\] where $f: 0,\ldots {k-1}$ is a filter of size $k$, $d$ a dilation factor and $s-d\cdot i$ accounts for the direction of the past. In this way, the context taken into account during the prediction can be increased by using larger filter sizes and by increasing the dilation factor. The most common practice is to increase the dilation factor exponentially with the depth of the network.

Recently,  \citet{devlin2018bert} proposed a novel approach to language modelling. Their BERT (Bidirectional Encoder Representations from Transformers) uses both left and right context, which means that a word $w\textsubscript{t}$ in a sequence is not determined just from its left sequence $w\textsubscript{1:{t-1}} = [w_1,...,w_{t-1}]$ but also from its right word sequence $w\textsubscript{t+1:n} = [w_{t+1},...,w_{t+n}]$. This approach introduces a new learning objective, a \textit{masked language model}, where a predefined percentage of randomly chosen words from the input word sequence is masked, and the objective is to predict these masked words from the unmasked context. BERT uses a transformer neural network architecture \citep{vaswani2017attention}, which relies on the self-attention mechanism. The distinguishing feature of this approach is the employment of several parallel attention layers, the so-called \textit{attention heads}, which reduce the computational cost and allow the system to attend to several dependencies at once.

All types of neural network language models, TCN, LSTM, and BERT, output softmax probability distribution calculated over the entire vocabulary, and present the probabilities for each word given its historical (and in case of BERT also future) sequence. Training of these networks usually minimizes the negative log-likelihood (NLL) of the training corpus word sequence $w\textsubscript{1:n} = [w_1,...,w_n]$ by backpropagation through time:
\begin{equation}
\textrm{NLL} = -\sum_{i=1}^{n}\log{P(w_i|w\textsubscript{1:i-1})}
\label{eq:NLL}
\end{equation}
In case of BERT, the formula for minimizing NLL uses also the right-hand  word sequence:

\[\textrm{NLL} = -\sum_{i=1}^{n}\log{P(w_i|w\textsubscript{1:i-1},  w\textsubscript{i+1:n})},\]where $w_i$ are the \textit{masked words}. 

The following equation, which is used for measuring the perplexity of neural language models, defines the relationship between perplexity (PPL, see Eq. (\ref{eq:PPL})) and NLL (Eq. (\ref{eq:NLL})):

\[\textrm{PPL} = e^{(\frac{\textrm{NLL}}{N})}\]

\subsection{Unsupervised methodology}
\label{sec-methodology-unsupervised}


Two main questions we wish to investigate in the unsupervised approach are the following:
\begin{itemize}
    \item Can standalone neural language models be used for unsupervised readability prediction?
    \item Can we develop a robust new readability formula that will outperform traditional readability formulas by relying not only on shallow lexical sophistication indicators but also on neural language model statistics?  
\end{itemize}

\subsubsection{Language models for unsupervised readability assessment}
\label{sec:lm_for_readability_assessment}

The findings of the related research suggest that a separate language model should be trained for each readability class in order to extract features for successful readability prediction \citep{petersen2009machine, xia2016text}. On the other hand, we test the possibility of using a neural language model as a standalone unsupervised readability predictor. 

First two points that support this kind of usage are based on the fact that neural language models tend to capture much more information compared to the traditional n-gram models. First, since n-gram language models used in the previous work on readability detection were in most cases limited to a small contextual window of up to five words, their learning potential was limited to lexico-semantic information (e.g., information about the difficulty of vocabulary and word n-gram structures in the text), and information about the text syntax. We argue that due to much larger contextual information of the neural models (e.g., BERT leverages sequences of up to 512 byte-pair tokens), which spans across sentences, the neural language models also learn high-level textual properties, such as long-distance dependencies \citep{jawahar-etal-2019-bert}, in order to minimize NLL during training. Secondly, n-gram models in the past readability research have only been trained on the corpora (or more specifically, on parts of the corpora) on which they were later employed. In contrast, by training the neural models on large general corpora, the model also learns semantic information, which can be transferred when the model is employed on a smaller test corpus. The success of this knowledge transfer is, to some extent, dependent on the genre compatibility of the train and test corpora. 

The third point favouring greater flexibility of neural language models relies on the fact that no corpus is a monolithic block of text made out of units (i.e. sentences, paragraphs, and articles) of exactly the same readability level. This means that a language model trained on a large corpus will be exposed to chunks of text with different levels of complexity. We hypothesize that due to this fact, the model will to some extent be able to distinguish between these levels and return a lower perplexity for more standard, predictable (i.e. readable) text. Vice versa, complex and rare language structures and vocabulary of less readable texts would negatively affect the performance of the language model, expressed via larger perplexity score. If this hypothesis is correct, ideally, the average readability of the training corpus should fit somewhere in the middle of the readability spectre of the testing corpus. 

To test the above statements, we train language models on Wiki-normal, Wiki-simple, and Wiki-balanced corpora described in Section \ref{sec:datasets}. All three Wiki corpora contain roughly the same amount of text, in order to make sure that the training set size does not influence the results of the experiments. 
 We expect the following results:
\begin{itemize}
    \item \textbf{Hypothesis 1:} Training the language models on a corpus with a readability that fits somewhere in the middle of the readability spectre of the testing corpus will yield the best correlation between the language model's performance and readability. According to the preliminary analysis of our corpora conducted in Section \ref{sec:dataset-analysis} and results of the analysis in Table \ref{table:datasets-stats}, this ideal scenario can be achieved in three cases: i) if a language model trained on the Wiki-simple is employed on the Newsela corpora, ii) if a language model trained on the Wiki-balanced corpus is employed on the OneStopEnglish corpus,  and iii) if the model trained on the KRES-balanced corpus is employed on the Slovenian SB corpus, despite the mismatch of genres in these corpora.
    \item \textbf{Hypothesis 2:} The language models trained only on texts for adults (Wiki-normal) will show higher perplexity on texts for children (WeeBit and Newsela) since their training set did not contain such texts; this will negatively affect the correlation between the language model's performance and readability.
    \item \textbf{Hypothesis 3:} Training the language models only on texts for children (Wiki-simple corpus) will result in a higher perplexity score of the language model when applied to adult texts (OneStopEnglish). This will positively affect the correlation between the language model's performance and readability. However, this language model will not be able to reliably distinguish between texts for different levels of adult ESL learners, which will have a negative effect on the correlation. 
\end{itemize}
 
To further test the viability of the unsupervised language models as readability predictors and to test the limits of using a single language model, we also explore the possibility of using a language model trained on a large general corpus. English BERT language model was trained on large corpora (Google Books Corpus \citep{goldberg2013dataset} and Wikipedia) of about 3300M words containing mostly texts for adult English speakers. According to hypothesis 2 above, this will have a negative effect on the correlation between the performance of the model and readability.

Due to the large size of the BERT's model and its huge training corpus, the semantic information acquired during training is much larger than the information acquired by the models we train on our much smaller corpora, which means that there is a greater possibility that the BERT model was trained on some text semantically similar to the content in the test corpora and that this information can be successfully transferred. However, the question remains, exactly what type of semantic content does the BERT's training corpus contain. One hypothesis is that its training corpus contains more content specific for adult audiences and less content found in the corpora for children. This would have a negative effect on the correlation between the performance of the model and readability on the WeeBit corpus. Contrarily, since the two highest readability classes in the WeeBit corpus contain articles from different scientific fields used for the education of high school students, which can contain rather specific and technical content that is unlikely to be common in the general training corpus, this might influence a positive correlation between the performance of the model and readability. The Newsela and OneStopEnglish, on the other hand, are parallel corpora, which means that the semantic content in all classes is very similar; therefore the success or failure of semantic transfer will most likely not affect these two corpora.

\subsubsection{Ranked sentence readability score}

Based on the two considerations below, we propose a new Ranked Sentence Readability Score (RSRS) for measuring the readability with language models.

\begin{itemize}
\item The shallow lexical sophistication indicators, such as the length of a sentence, correlate well with the readability of a text. Using them besides statistics derived from language models could improve the unsupervised readability prediction. 
\item The perplexity score used for measuring the performance of a language model is an \emph{unweighted} sum of perplexities of words in the predicted sequence. In reality, a small number of unreadable words might drastically reduce the readability of the entire text. Assigning larger weights to such words might improve the correlation of language model scores with the readability.
\end{itemize}

\begin{figure}[t!]
  \centering
  \includegraphics[scale=0.43]{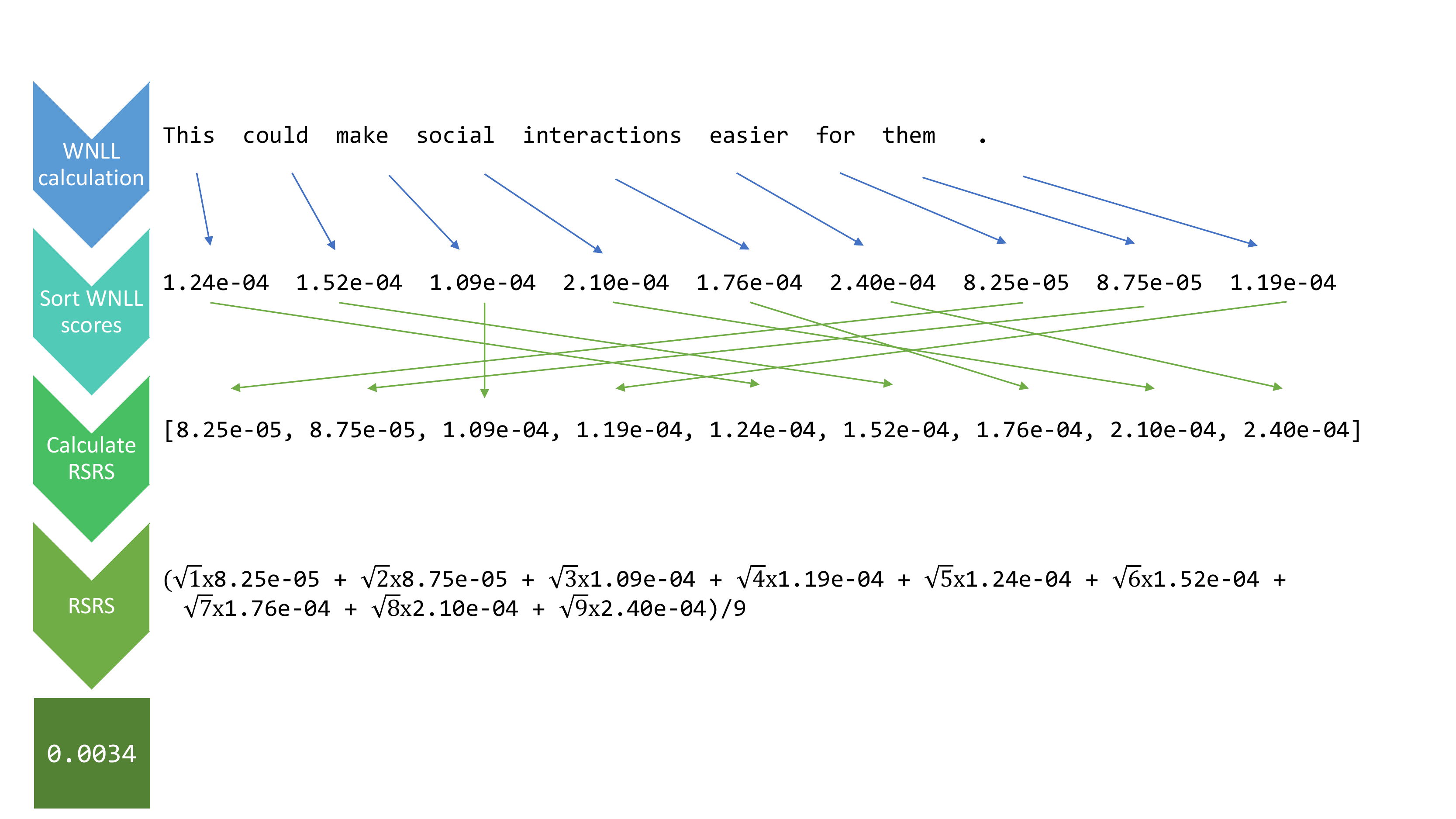}
  \caption{The RSRS calculation for the sentence \textit{This could make social interactions easier for them.}}
  \label{rsrs-example}
\end{figure}

The proposed readability score is calculated with the following procedure. First, a given text is split into sentences with the default sentence tokenizer from the NLTK library \citep{bird2004nltk}. In order to get a readability estimation for each word in a specific context, we compute, for each word in the sentence, the word negative log-likelihood (WNLL) according to the following formula:

\[\textrm{WNLL}= -(y_t \log{y_p} + (1 - y_t) \log{(1 - y_p)}),\]
\noindent
where $y_p$ denotes the probability (from the softmax distribution) predicted by the language model according to the historical sequence, and $y_t$ denotes the empirical distribution for a specific position in the sentence, i.e. $y_t$ has the value 1 for the word in the vocabulary that actually appears next in the sequence and the value 0 for all the other words in the vocabulary. Next, we sort all the words in the sentence in ascending order according to their WNLL score, and the ranked sentence readability score (RSRS) is calculated with the following expression:
\begin{equation}
 \textrm{RSRS} = \frac{\sum_{i = 1}^{S} \sqrt{i} \cdot \textrm{WNLL}(i)}{S},
 \label{eq:RSRS}
\end{equation}
\noindent
where $S$ denotes the sentence length and $i$ represents the rank of a word in a sentence according to its WNLL value. The square root of the word rank is used for proportionally weighting words according to their readability since initial experiments suggested that the use of a square root of a rank represents the best balance between allowing all words to contribute equally to the overall readability of the sentence and allowing only the least readable words to affect the overall readability of the sentence. For out of vocabulary words, square root rank weights are doubled, since these rare words are, in our opinion, good indicators of non-standard text. Finally, in order to get the readability score for the entire text, we calculate the average of all the RSRS scores in the text. An example of how RSRS is calculated for a specific sentence is shown in Figure \ref{rsrs-example}.

The main idea behind the RSRS score is to avoid the reductionism of traditional readability formulas. We aim to achieve this by including high-level structural and semantic information through neural language model based statistics. The first assumption is that complex grammatical and lexical structures harm the performance of the language model. Since WNLL score, which we compute for each word, depends on the context in which the word appears in, words appearing in more complex grammatical and lexical contexts will have a higher WNLL. The second assumption is that the semantic information is included in the readability calculation: tested documents with semantics dissimilar to the documents in the language model training set will negatively affect the performance of the language model, resulting in the higher WNLL score for words with unknown semantics. The trainable nature of language models allows for customization and personalization of the RSRS for specific tasks, topics and languages. This means that RSRS shall alleviate the problem of cultural non-transferability of traditional readability formulas.   

On the other hand, the RSRS also leverages shallow lexical sophistication indicators through the index weighting scheme, which makes sure that less readable words contribute more to the overall readability score. This is somewhat similar to the counts of long and difficult words in the traditional readability formulas, such as GFI and DCRF. The value of RSRS also increases for texts containing longer sentences, since the square roots of the word rank weights become larger with increased sentence length. This is similar to the behaviour of traditional formulas such as GFI, FRE, FKGL, ARI, and DCRF, where this effect is achieved by incorporating the ratio between the total number of words and the total number of sentences into the equation.  

\subsection{Unsupervised experiments}
\label{sec-unsupervised-experiments}
For the presented unsupervised readability assessment methodology based on neural language models, we first present the experimental design followed by the results.

\subsubsection{Experimental design}
Three different architectures of language models (described in Section \ref{sec:NLMs}) are used for experiments: a temporal convolutional network (TCN) proposed by  \citet{bai2018empirical}, a recurrent language model (RLM) using character-level CNN and LSTM proposed by \citet{kim2016character}, and an attention-based language model BERT \citep{devlin2018bert}. For the experiments on the English language, we train TCN and RLM on three Wiki corpora.

To explore the possibility of using a language model trained on a general corpus for the unsupervised readability prediction, we use the bert-base-uncased English language model, a pretrained uncased language model trained on BooksCorpus (0.8G words) \citep{zhu2015aligning} and English Wikipedia (2.5G words). For the experiments on Slovenian language, the corpus containing just school books is too small for efficient training of language models; therefore TCN and RLM were only trained on the KRES-balanced corpus described in Section \ref{sec:datasets}. For exploring the possibility of using a general language model for the unsupervised readability prediction, a pretrained CroSloEngual BERT model trained on corpora from three languages, Slovenian (1.26G words), Croatian (1.95G words), and English (2.69G words) \citep{ulcar2020xlbert}, is used. The corpora used in training of the model are a mix of news articles and a general web crawl.  

The performance of language models is typically measured with the perplexity (see Eq. (\ref{eq:PPL})). To answer the research question if standalone language models can be used for unsupervised readability prediction, we investigate how the measured perplexity of language models correlates with the readability labels in the gold-standard WeeBit, OneStopEnglish, Newsela, and Slovenian SB corpora described in Section \ref{sec:datasets}. The correlation to these ground truth readability labels is also used to evaluate the performance of the RSRS measure. For performance comparison, we calculate the traditional readability formula values (described in Section \ref{related-work}) for each document in the gold-standard corpora and measure the correlation between these values and manually assigned labels. As a baseline, we use the average sentence length (ASL) in each document.

The correlation is measured with the Pearson correlation coefficient ($\rho$). Given a pair of distributions $X$ and $Y$, the covariance $cov$, and the standard deviation $\sigma$, the formula for $\rho$ is:

\[ \rho_{x,y} = \frac{cov(x,y)}{\sigma_x  \sigma_y} \]

A larger positive correlation signifies a better performance for all measures except the FRE readability measure. As this formula assigns higher scores to better readable texts, a larger negative correlation suggests a better performance of the FRE measure. 

\subsubsection{Experimental results}
\label{sec-results-unsupervised}

The results of the experiments are presented in Table \ref{table:resultsUnsupervised}. The ranking of measures on English and Slovenian datasets are presented in Table \ref{table:rankEng}. 

\begin{table*}[b]
\begin{center}
\caption{Pearson correlation coefficient between manually assigned readability labels and the readability scores assigned by different readability measures in the unsupervised setting. The highest correlation for each corpus is marked with the bold typeface.} 
\begin{tabular}{lrrrr}
  Measure/Dataset & WeeBit & OneStopEnglish & Newsela & Slovenian SB \\
  \hline     
  RLM perplexity-balanced & -0.082 & 0.405 & 0.512 & 0.303 \\
  RLM perplexity-simple & -0.115 & 0.420 & 0.470 & / \\
  RLM perplexity-normal & -0.127 & 0.283 & 0.341 & / \\
  TCN perplexity-balanced & 0.034 & 0.476 & 0.537 & 0.173 \\
  TCN perplexity-simple & 0.025 & 0.518 & 0.566 & / \\
  TCN perplexity-normal & -0.015 & 0.303 & 0.250 & / \\
  BERT perplexity & -0.123 & -0.162 & -0.673 & -0.563  \\
  \hline
  RLM RSRS-balanced & 0.497 & 0.551 & 0.890 & 0.732 \\
  RLM RSRS-simple & 0.506 & 0.569 & 0.893 & / \\
  RLM RSRS-normal & 0.490 & 0.536 & 0.886 & / \\
  TCN RSRS-balanced & 0.393 & 0.601 & 0.894 &  \textbf{0.789} \\
  TCN RSRS-simple & 0.385 & \textbf{0.615} & 0.894 & / \\
  TCN RSRS-normal & 0.348 & 0.582 & 0.886 & / \\
  BERT RSRS & 0.279 & 0.384 & 0.674 & 0.126 \\
  \hline
  GFI & \textbf{0.544} & 0.550 & 0.849 & 0.730 \\
  FRE & -0.433 & -0.485 & -0.775 & -0.614 \\
  FKGL & \textbf{0.544} & 0.533 & 0.865 & 0.697\\
  ARI & 0.488 & 0.520 & 0.875 & 0.658 \\
  DCRF & 0.420 & 0.496 & 0.735 & 0.686 \\
  SMOG & 0.456 & 0.498 & 0.813 & 0.770 \\
  ASL & 0.508 & 0.498 & \textbf{0.906} & 0.683 \\
  \hline
\end{tabular}
\label{table:resultsUnsupervised}
 \end{center}
\end{table*}

\begin{table*}[t]
\begin{center}
\caption{Ranking (lower is better) of measures on English and Slovenian datasets sorted by the average rank on all datasets for which the measure is available.}
\begin{tabular}{lrrrrr}
    \hline
    Measure & WeeBit & OneStopEnglish & Newsela & Slovenian SB \\
    RLM RSRS-simple & 4 & 4 & 4 & /  \\
    TCN RSRS-balanced & 11 & 2 & 2 & \textbf{1}  \\
    RLM RSRS-balanced & 5 & 5 & 5 & 3  \\
    GFI & \textbf{1} & 6 & 10 & 4  \\
    TCN RSRS-simple & 12 & \textbf{1} & 3 & /  \\
    ASL & 3 & 12 & \textbf{1} & 7  \\
    FKGL & 2 & 8 & 9 & 5  \\
    RLM RSRS-normal & 6 & 7 & 6 & /  \\
    TCN RSRS-normal & 13 & 3 & 7 & /  \\
    ARI & 7 & 9 & 8 & 8  \\
    SMOG & 8 & 11 & 11 & 2  \\
    DCRF & 10 & 13 & 13 & 6  \\
    FRE & 9 & 14 & 12 & 9  \\
    TCN perplexity-simple & 16 & 10 & 15 & / \\
    TCN perplexity-balanced & 15 & 15 & 16 & 11 \\
    BERT RSRS & 14 & 18 & 14 & 12  \\
    RLM perplexity-balanced & 18 & 17 & 17 & 10  \\
    RLM perplexity-simple & 19 & 16 & 18 & / \\
    TCN perplexity-normal & 17 & 19 & 20 & /  \\
    BERT perplexity & 20 & 21 & 21 & 13  \\
    RLM perplexity-normal & 21 & 20 & 19 & / \\
    \hline
\end{tabular}
\label{table:rankEng}
 \end{center}
\end{table*}

The correlation coefficients of all measures vary drastically between different corpora. The highest $\rho$ values are obtained on the Newsela corpus, where the best performing measure (surprisingly this is our baseline - the average sentence length) achieves the $\rho$ of 0.906. The highest $\rho$ on the other two English corpora are much lower. On the WeeBit corpus, the best performance is achieved by GFI and FKGL measures ($\rho$ of 0.544), and on the OneStopEnglish corpus, the best performance is achieved with the proposed TCN RSRS-simple ($\rho$ of 0.615). On the Slovenian SB, the $\rho$ values are higher, and the best performing measure is TCN RSRS score-balanced with $\rho$ of 0.789.

The perplexity-based measures show a much lower correlation with the ground truth readability scores. Overall, they perform the worst of all the measures for both languages (see Table \ref{table:rankEng}), but we can observe large differences in their performance across different corpora. While there is either no correlation or low negative correlation between perplexities of all three language models and readability on the WeeBit corpus, there is some correlation between perplexities achieved by RLM and TCN on OneStopEnglish and Newsela corpora (the highest being the $\rho$ of 0.566 achieved by TCN perplexity-simple on the Newsela corpus). The correlation between RLM and TCN perplexity measures and readability classes on the Slovenian SB corpus is low, with RLM perplexity-balanced showing the $\rho$ of 0.303 and TCN perplexity-balanced achieving $\rho$ of 0.173.  

BERT perplexities are negatively correlated with readability, and the negative correlation is relatively strong on Newsela and Slovenian school books corpora ($\rho$ of \num{-0.673} and \num{-0.563}, respectively), and weak on WeeBit and OneStopEnglish corpora. As BERT was trained on corpora which are mostly aimed at adults, the strong negative correlation on Newsela and Slovenian SB corpora seem to suggest that BERT language models might actually be less perplexed by the articles aimed at adults than the documents aimed at younger audiences. This is supported by the fact that the negative correlation is weaker on the OneStopEnglish corpus, which is meant for adult audiences, and for which our analysis (see Section \ref{sec:dataset-analysis}) has shown that it contains more complex texts according to the shallow readability indicators. 

Nevertheless, the weak negative correlation on the WeeBit corpus is difficult to explain as one would expect a stronger negative correlation because the same analysis showed that WeeBit contains least complex texts out of all the tested corpora. If this result is connected with the successful transfer of the semantic knowledge, it supports the hypothesis that the two classes containing most complex texts in the WeeBit corpus contain articles with rather technical content that perplex the BERT model. However, the role of the semantic transfer should also dampen the negative correlation on the Slovenian SB, which is a non-parallel corpus and also contains rather technical educational content meant for high-school children. Perhaps the transfer is less successful for Slovenian since the Slovenian corpus on which the CroSloEngual BERT was trained is smaller than the English corpora used for training of English BERT. While further experiments and data are needed to pinpoint the exact causes for the discrepancies in the results, we can still conclude that using a single language model trained on general corpora for unsupervised readability prediction of texts for younger audiences or English learners is, at least according to our results, not a viable option.

Regarding our expectations that performance of the language model trained on a corpus with average readability that fits somewhere in the middle of the readability spectre of the testing corpus would yield the best correlation with manually labelled readability scores, it is interesting to look at the differences in performance between TCN and RLM perplexity measures trained on Wiki-normal, Wiki-simple and Wiki-balanced corpora. As expected, the correlation scores are worse on the WeeBit corpus, since all classes in this corpus contain texts that are less complex than texts in any of the training corpora. On the OneStopEnglish corpus, both Wiki-simple perplexity measures perform the best, which is unexpected, since we would expect the balanced measure to perform better. On the Newsela corpus, RLM perplexity-balanced outperforms RLM perplexity-simple by 0.042 (which is unexpected), and TCN perplexity-simple outperforms TCN perplexity-balanced by 0.029, which is according to the expectations. Also, according to the expectation is the fact, that both Wiki-normal perplexity measures are outperformed by a large margin by Wiki-simple and Wiki-balanced perplexity measures on the OneStopEnglish and the Newsela corpora. Similar observations can be made in regards to RSRS, which also leverages language model statistics. On all corpora, the performance of Wiki-simple RSRS measures and Wiki-balanced RSRS measures is comparable, and these measures consistently outperform Wiki-normal RSRS measures. 

These results are not entirely compatible with hypothesis 1 in Section \ref{sec:lm_for_readability_assessment} that Wiki-balanced measures would be most correlated with readability on the OneStopEnglish corpus and that Wiki-simple measures would be most correlated with readability on the Newsela corpus. Nevertheless, training the language models on the corpora with readability in the middle of the readability spectre of the test corpus seems to be an effective strategy, since the differences in performance between Wiki-balanced and Wiki-simple measures are not large. On the other hand, the good performance of the Wiki-simple measures supports our hypothesis 3 in Section \ref{sec:lm_for_readability_assessment}, that training the language models on texts with the readability closer to the bottom of the readability spectrum of the test corpus for children will result in a higher perplexity score of the language model when applied to adult texts, which will have a positive effect on the correlation with readability. 

The fact that positive correlation between readability and both Wiki-simple and Wiki-balanced perplexity measures on the Newsela and OneStopEnglish corpora is quite strong supports the hypothesis that more complex language structures and vocabularies of less readable texts would result in a higher perplexity on these texts. Interestingly, strong correlations also indicate that the genre discrepancies between the language model train and test sets do not appear to have a strong influence on the performance. While the choice of a neural architecture for language modelling does not appear to be that crucial, the readability of the language model training set is of utmost importance. If the training set on average contains more complex texts than the majority of texts in the test set, as in the case of language models trained just on the Wiki-normal corpus (and also BERTs), the correlation between readability and perplexity disappears or even gets reverted, since language models trained on more complex language structures learn how to handle these difficulties. 

The low performance of perplexity measures suggests that neural language model statistics are not good indicators of readability and should therefore not be used alone for readability prediction. Nevertheless, the results of TCN RSRS and RLM RSRS suggest that language models contain quite useful information when combined with other shallow lexical sophistication indicators, especially when readability analysis needs to be conducted on a variety of different datasets. 

As seen in Table \ref{table:rankEng}, shallow readability predictors can give inconsistent results on datasets from different genres and languages. For example, the simplest readability measure, the average sentence length, ranked first on Newsela and twelfth on OneStopEnglish. It also did not do well on the Slovenian SB corpus, where it ranked seventh. SMOG, on the other hand, ranked very well on the Slovenian SB corpus (rank 2) but ranked twice as eleventh and once as eighth on the English corpora. Among the traditional measures, GFI presents the best balance in performance and consistency, ranking first on WeeBit, sixth on OneStopEnglish, tenth on Newsela, and fourth on Slovenian SB.

On the other hand, RSRS-simple and RSRS-balanced measures offer more robust performance across datasets from different genres and languages according to ranks in Table \ref{table:rankEng}. For example, the RLM RSRS-simple measure ranked fourth on all English corpora. The TCN RSRS-balanced measure, which was also employed on Slovenian SB, ranked first on Slovenian SB and second on OneStopEnglish and Newsela. However, it did not do well on WeeBit, where the discrepancy in readability between the language model train and test sets was too large. RLM RSRS-balanced was more consistent, ranking fifth on all English corpora and third on Slovenian SB. These results suggest that language model statistics can improve the consistency of predictions on a variety of different datasets. The robustness of the measure is achieved by training the language model on a specific train set, with which one can optimize the RSRS measure for a specific task and language.

\section{Supervised neural approach}
\label{sec:supervised}

As mentioned in Section \ref{sec-intro}, recent trends in text classification show the domination of deep learning approaches which internally employ automatic feature construction. Existing neural approaches to readability prediction (see Section \ref{sec:neural-classification}) tend to generalize better across datasets and genres \cite{filighera2019automatic}, and therefore solve the problem of classical machine learning approaches relying on an extensive feature engineering \citep{xia2016text}. 

In this section, we analyze how different types of neural classifiers can predict text readability. In Section \ref{sec-methodology-supervised}, we describe the methodology, and in Section \ref{sec-results-supervised} we present experimental scenarios and results of conducted experiments.

\subsection{Supervised methodology}
\label{sec-methodology-supervised}
We tested three distinct neural network approaches to text classification:
\begin{itemize}
    \item Bidirectional Long short-term memory network (BiLSTM). We use the RNN approach proposed by \citet{conneau2017supervised} for classification. The bidirectional LSTM layer is a concatenation of forward and backward LSTM layers that read documents in two opposite directions. The max and mean pooling are applied to the LSTM output feature matrix in order to get the maximum and average values of the matrix. The resulting vectors are concatenated and fed to the final linear layer responsible for predictions.
    \item Hierarchical attention networks (HAN). We use the architecture of \citet{yang2016hierarchical} that takes hierarchical structure of text into account with the two-level attention mechanism \citep{bahdanau2014neural, xu2015show} applied to word and sentence representations encoded by bidirectional LSTMs.
    \item Transfer learning. We use the pretrained BERT transformer architecture with 12 layers of size 768 and 12 self-attention heads. A linear classification head was added on top of the pretrained language model, and the whole classification model was fine-tuned on every dataset for 3 epochs. For English datasets, the bert-base-uncased English language model is used, while for the Slovenian SB corpus, we use the CroSloEngual BERT model trained on Slovenian, Croatian and English \citep{ulcar2020xlbert}\footnote{Both models are available through the Transformers library \url{https://huggingface.co/transformers/}.}. 
    
\end{itemize}

We randomly shuffle all the corpora, and then Newsela and Slovenian SB corpora are split into a train (80\% of the corpus), validation (10\% of the corpus) and test (10\% of the corpus) sets. Due to the small number of documents in OneStopEnglish and WeeBit corpora (see description in Section \ref{sec:datasets}), we used five-fold stratified cross-validation on these corpora to get more reliable results. For every fold, the corpora were split into the train (80\% of the corpus), validation (10\% of the corpus) and test (10\% of the corpus) sets. We employ Scikit StratifiedKFold\footnote{\url{https://scikit-learn.org/stable/modules/generated/sklearn.model_selection.StratifiedKFold.html}}, both for train-test splits and five-fold cross-validation splits, in order to preserve the percentage of samples from each class.

BiLSTM and HAN classifiers were trained on the train set and tested on the validation set after every epoch (for a maximum of 100 epochs). The best performing model on the validation set was selected as the final model and produced predictions on the test sets. BERT models are fine-tuned on the train set for three epochs, and the resulting model is tested on the test set. The validation sets were used in a grid search to find the best hyperparameters of the models. For BiLSTM, all combinations of the following hyperparameter values were tested before choosing the best combination, which is written in bold in the list below:

\begin{itemize}
\item Batch size: \textbf{8}, 16, 32
\item Learning rates: 0.00005, \textbf{0.0001}, 0.0002, 0.0004, 0.0008
\item Word embedding size: 100, \textbf{200}, 400
\item LSTM layer size: \textbf{128}, 256
\item Number of LSTM layers: 1, \textbf{2}, 3, 4
\item Dropout after every LSTM layer: 0.2,  \textbf{0.3}, 0.4
\end{itemize}

For HAN, we tested all combinations of the following hyperparameter values (the best combination is written in bold):

\begin{itemize}
\item Batch size: 8, \textbf{16}, 32
\item Learning rates: 0.00005, \textbf{0.0001}, 0.0002, 0.0004, 0.0008
\item Word embedding size: 100, \textbf{200}, 400
\item Sentence embedding size: \textbf{100}, 200, 400
\end{itemize}

For BERT fine-tuning, we use the default learning rate of 0.00002. The input sequence length is limited to 512 byte-pair tokens, which is the maximum supported input sequence length.

We used the same configuration for all the corpora and performed no corpus specific tweaking of classifier parameters. We measured the performance of all the classifiers in terms of accuracy (in order to compare their performance to the performance of the classifiers from the related work), weighted average precision, weighted average recall, and weighted average $F_1$-score\footnote{We employ the Scikit implementation of the metrics (\url{https://scikit-learn.org/stable/modules/classes.html\#module-sklearn.metrics}) and set the ``average'' parameter to ``weighted''.}. Since readability classes are ordinal variables (in our case ranging from 0 to n=number of classes-1), not all mistakes of classifiers are equal; therefore we also employ the Quadratic Weighted Kappa (QWK) measure, which allows for mispredictions to be weighted differently, according to the cost of a specific mistake. Calculation of the QWK involves three matrices containing observed scores, ground truth scores and the weight matrix scores, which in our case correspond to the distance $d$ between the classes $c_i$ and $c_j$ and is defined as $d=|c_i-c_j|$. QWK is therefore calculated as:

\begin{equation}
\textrm{QWK} = 1- \frac{\sum_{i=1}^{c} \sum_{j=1}^{c}w_{ij}x_{ij}}{\sum_{i=1}^{c} \sum_{j=1}^{c}w_{ij}m_{ij}},
\label{eq:QWKL}
\end{equation}
\noindent 

\noindent where $c$ is the number of readability classes and $w_{ij}$, $x_{ij}$ and $m_{ij}$ are elements in the weight, observed, and ground truth matrices, respectively.

\subsection{Supervised experimental results}
\label{sec-results-supervised}

The results of supervised readability assessment using different architectures of deep neural networks are presented in Table \ref{table:supervisedResults} together with the state-of-the-art baseline results from the related work \citep{filighera2019automatic, xia2016text, deutsch2020linguistic}. We only present the best result reported by each of the baseline studies, the only exception is \citet{deutsch2020linguistic}, for which we present two results, SVM-BF (SVM with BERT features) and SVM-HF (SVM with HAN features) that proved the best on the WeeBit and Newsela corpora, respectively.

\begin{table*}[b]
\begin{center}
\caption{The results of the supervised approach to readability in terms of accuracy, weighted precision, weighted recall, and weighted $F_1$-score for the three neural network classifiers and  methods from the literature.} 
\begin{tabular}{lcccc}
  Measure/Dataset & WeeBit & OneStopEnglish & Newsela & Slovenian SB\\
  \hline
  Filighera et al. (2019) accuracy & 0.8130 & - & - & -  \\
  Xia et al. (2016) accuracy & 0.8030 & - & - & -  \\
  SVM-BF (Deutsh et al., 2020) $F_1$ & 0.8381 & - & 0.7627 & -  \\
  SVM-HF (Deutsh et al., 2020) $F_1$ &  - & - & 0.8014 & -  \\
  Vajjala et al. (2018) accuracy & - & 0.7813 & - & - \\
  
  \hline     
    BERT accuracy & \textbf{0.8573} & 0.6738 & 0.7573 & 0.4545 \\
    BERT precision & \textbf{0.8658} & 0.7395 & 0.7510 & 0.4736 \\
    BERT recall & \textbf{0.8573} & 0.6738 & 0.7573 & 0.4545 \\
    BERT $F_1$ & \textbf{0.8581} & 0.6772 & 0.7514 & 0.4157 \\
    BERT QWK & \textbf{0.9527} & 0.7077 & 0.9789 & \textbf{0.8855} \\
  \hline
    HAN accuracy & 0.7520 & \textbf{0.7872} & \textbf{0.8138} &  0.4887 \\
    HAN precision & 0.7534 & \textbf{0.7977} & \textbf{ 0.8147} & 0.4866 \\
    HAN recall & 0.7520 & \textbf{0.7872} & \textbf{0.8138} & 0.4887 \\
    HAN $F_1$ & 0.7520 & \textbf{0.7888} & \textbf{0.8101} & 0.4847 \\
    HAN QWK & 0.8860 & \textbf{0.8245} & \textbf{0.9835} &0.8070 \\
  \hline
    BiLSTM accuracy & 0.7743 & 0.6875 & 0.7111 & \textbf{0.5277}\\
    BiLSTM precision & 0.7802 & 0.7177 & 0.6910 & \textbf{0.5239}\\
    BiLSTM recall & 0.7743 & 0.6875 & 0.7111 &  \textbf{0.5277}\\
    BiLSTM $F_1$ & 0.7750 & 0.6920 & 0.6985 & \textbf{0.5219}\\
    BiLSTM QWK & 0.9060 & 0.7230 & 0.9628 & 0.7980\\
  \hline
\end{tabular}
\label{table:supervisedResults}
 \end{center}
\end{table*}

On the WeeBit corpus, by far the best performance according to all measures was achieved by BERT. In terms of accuracy, BERT  outperforms the second-best BiLSTM by about 8 percentage points, achieving the accuracy of 85.73\%. HAN performs the worst on the WeeBit corpus according to all measures. BERT also outperforms the accuracy result reported by \citet{xia2016text}, who used the five-fold cross-validation setting and the accuracy result on the development set reported by \citet{filighera2019automatic}\footnote{For the study by \citet{filighera2019automatic}, we report accuracy on the development set instead of the accuracy on the test set, as the authors claim that this result is more comparable to the results achieved in the cross-validation setting. On the test set, \citet{filighera2019automatic} report the best accuracy of 74,4\%.}. In terms of weighted $F_1$-score, both strategies that employ BERT (employing the BERT classifier directly or feeding BERT features to the SVM classifier as in \citet{deutsch2020linguistic}) seem to return similar results. Finally, in terms of QWK, BERT achieves a very high score of 95.27\% and the other two tested classifiers obtain a good   QWK score close to 90\%.  

The best result on Newsela is achieved by HAN, achieving the $F_1$-score of 81.01\% and accuracy of 81.38\%. This is similar to the baseline SVM-HF result achieved by \citet{deutsch2020linguistic}, who fed HAN features to the SVM classifier. BERT performs less competitively on the OneStopEnglish and Newsela corpora. On OneStopEnglish, it is outperformed by the best performing classifier (HAN) by about 10 percentage points, and on Newsela, it is outperformed by about 6 percentage points according to accuracy and $F_1$ criteria. The most likely reason for the bad performance of BERT on these two corpora is the length of documents in these two datasets. On average, documents in the OneStopEnglish and Newsela corpora are 677 and 760 words long. On the other hand, BERT only allows input documents of up to 512 byte-pair tokens, which means that documents longer than that need to be truncated. This results in the substantial loss of information on the OneStopEnglish and Newsela corpora but not on the WeeBit and Slovenian SB corpora, which contain shorter documents, 193 and 305 words long. 

The results show that BiLSTM also has problems when dealing with longer texts, even though it does not require input truncation. This suggests that the loss of context is not the only reason for the non-competitive performance of BERT and BiLSTM, and that the key to the successful classification of long documents is the leveraging of hierarchical information in the documents, for which HAN was built for. The assumption is that this is particularly important in parallel corpora, where the simplified versions of the original texts contain the same message as the original texts, which forces the classifiers not to rely as much on semantic differences but rather focus on structural differences. 

While $F_1$-scores and accuracies suggest large discrepancies in performance between HAN and other two classifiers on the OneStopEnglish and Newsela corpora, QWK scores draw a different picture. While the discrepancy is still large on OneStopEnglish, all classifiers achieve almost perfect QWK scores on the Newsela dataset. This suggests that even though BERT and BiLSTM make more classification mistakes than HAN, these mistakes are seldom costly on the Newsela corpus (i.e. documents are classified into neighbouring classes of the correct readability class). QWK scores achieved on the Newsela corpus by all classifiers are also much higher than the scores achieved on other corpora (except for the QWK score achieved by BERT on the WeeBit corpus). This is in line with the results in the unsupervised setting, where the $\rho$ values on the Newsela corpus were substantially larger than on other corpora.    

The HAN classifier achieves the best performance on the OneStopEnglish corpus with the accuracy of 78.72\% in the five-fold cross-validation setting. This is comparable to the state-of-the-art accuracy of 78.13\% achieved by \citet{vajjala2018onestopenglish} with their SMO classifier using 155 hand-crafted features. BiLSTM and BERT classifiers perform similarly on this corpus, by about 10 percentage points worse than HAN, according to accuracy, $F_1$-score, and QWK. 

The results on the  Slovenian SB corpus are also interesting. In general, the performance of classifiers is the worst on this corpus, with the $F_1$-score of 52.19\% achieved by BiLSTM being the best result. BiLSTM performs by about 4 percentage points better than HAN according to $F_1$-score and accuracy, while both classifiers achieve roughly the same QWK score of about 80\%. On the other hand, BERT achieves lower $F_1$-score (about 45.45\%) and accuracy (41.57\%), but performs much better than the other two classifiers according to QWK, achieving QWK of almost 90\%. 

\begin{figure}[t]
\minipage{0.33\textwidth}
  \includegraphics[width=1.2 \linewidth]{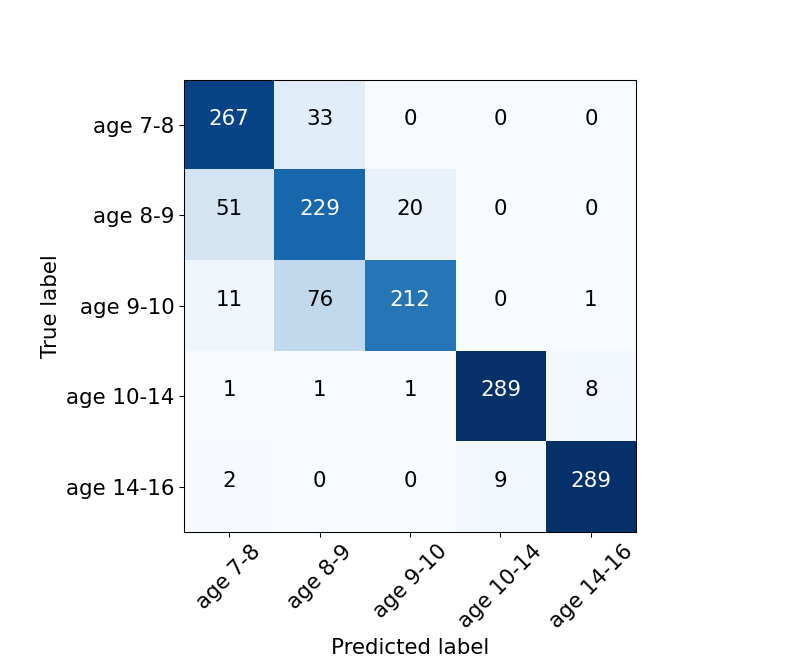}
  \centerline{a) BERT}
\endminipage\hfill
\minipage{0.33\textwidth}
  \includegraphics[width=1.2 \linewidth]{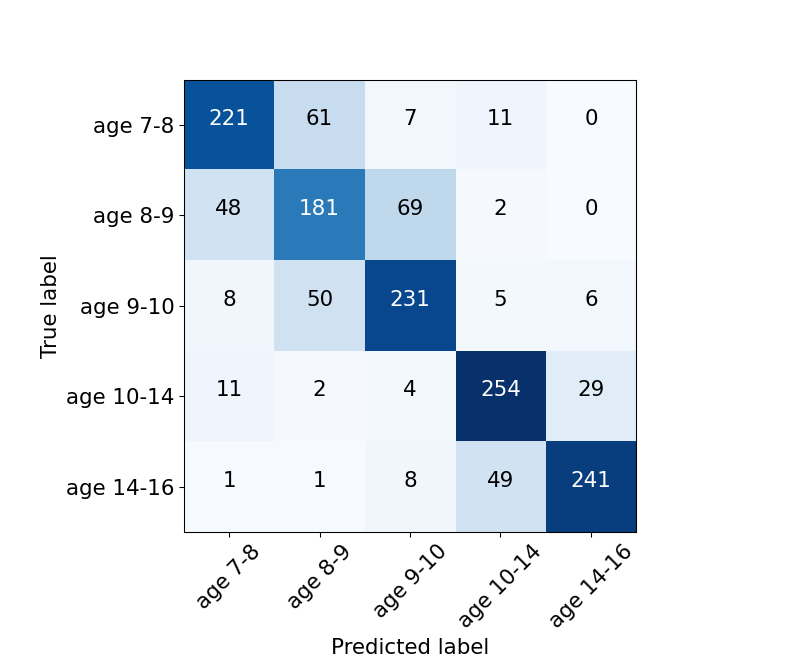}
    \centerline{b) HAN}
\endminipage\hfill
%
\minipage{0.33\textwidth}
  \includegraphics[width=1.2 \linewidth]{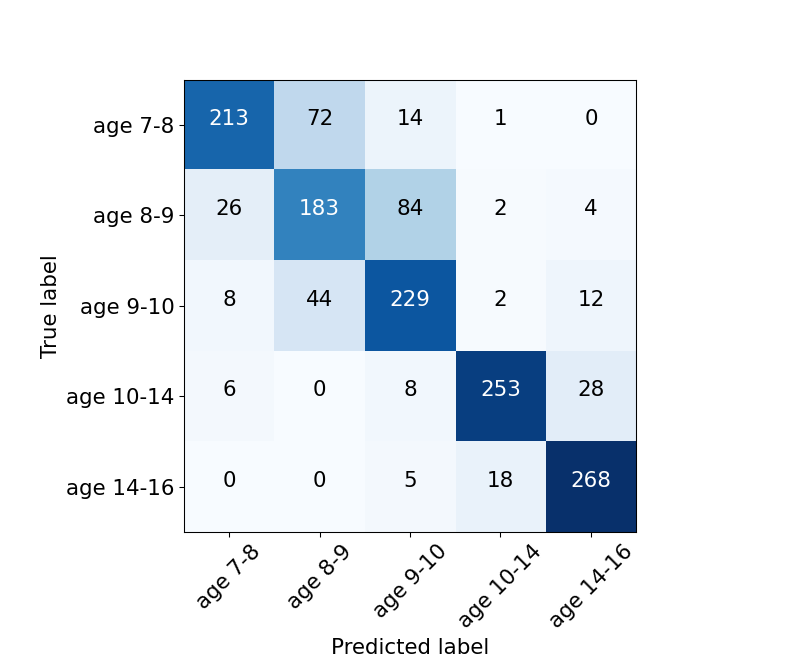}
    \centerline{c) BiLSTM}
\endminipage
 \caption{Confusion matrices for BERT, HAN, and BiLSTM on the WeeBit corpus.} 
   \label{fig:cm_weebit}
\end{figure}

Confusion matrices for classifiers give us a better insight into what kind of mistakes are specific for different classifiers. For the WeeBit corpus confusion matrices show (Figure \ref{fig:cm_weebit}) that all the tested classifiers have the most problems distinguishing between texts for children 8-9 years old and 9-10 years old. The mistakes where the text is falsely classified into an age group that is not neighbouring the correct age group are rare. For example, the best performing BERT classifier misclassified only sixteen documents into non-neighbouring classes. When it comes to distinguishing between neighbouring classes, the easiest distinction for the classifiers was the distinction between texts for children 9-10 years old and 10-14 years old. Besides fitting into two distinct age groups, the documents in these two classes also belong to two different sources (texts for children 9-10 years old consist of articles from WeeklyReader and texts for children 10-14 years old consist of articles from BBC-Bitesize), which suggests that the semantic and writing style dissimilarities between these two neighbouring classes might be larger than for other neighbouring classes, and that might have a positive effect on the performance of the classifiers.

\begin{figure}[t]
\minipage{0.33\textwidth}
  \includegraphics[width=1.2 \linewidth]{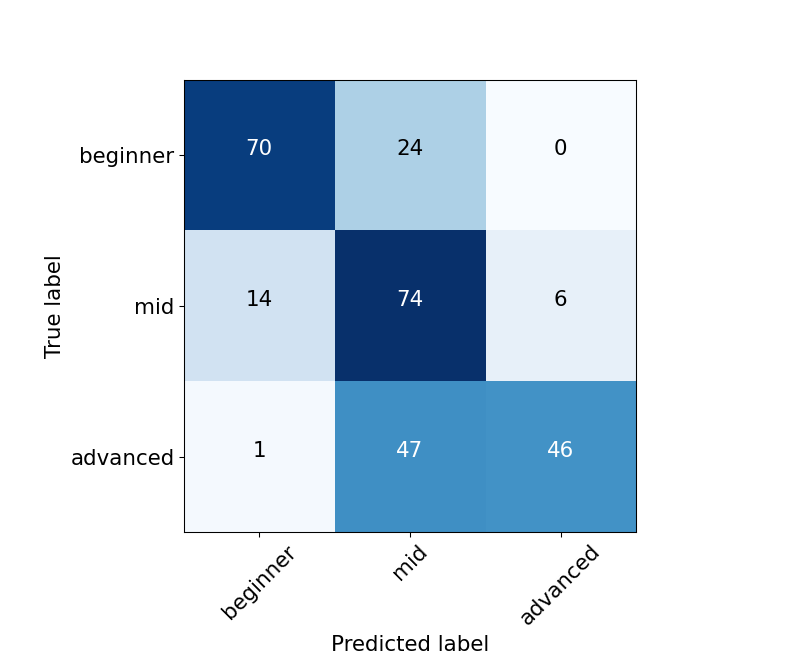}
  \centerline{a) BERT}
\endminipage\hfill
\minipage{0.33\textwidth}
  \includegraphics[width=1.2 \linewidth]{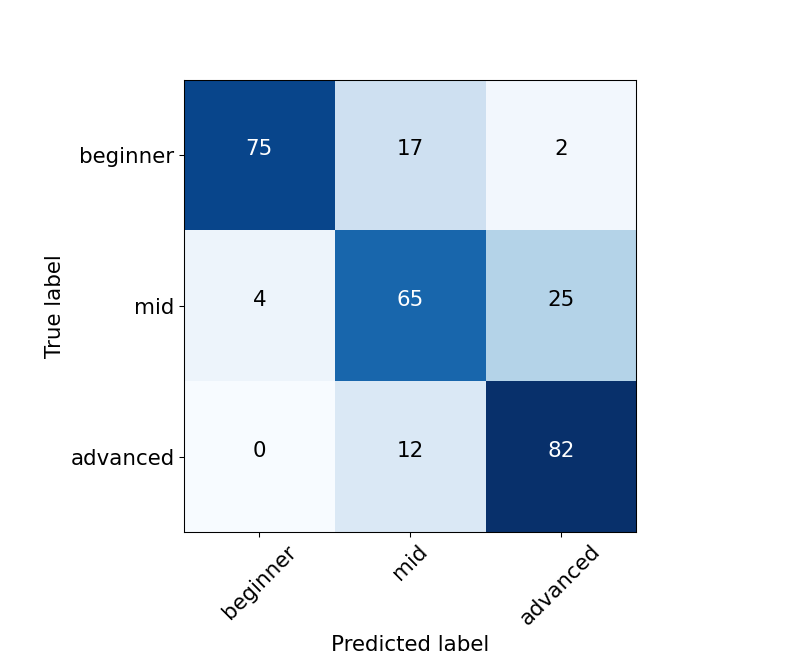}
    \centerline{b) HAN}
\endminipage\hfill
\minipage{0.33\textwidth}
  \includegraphics[width=1.2 \linewidth]{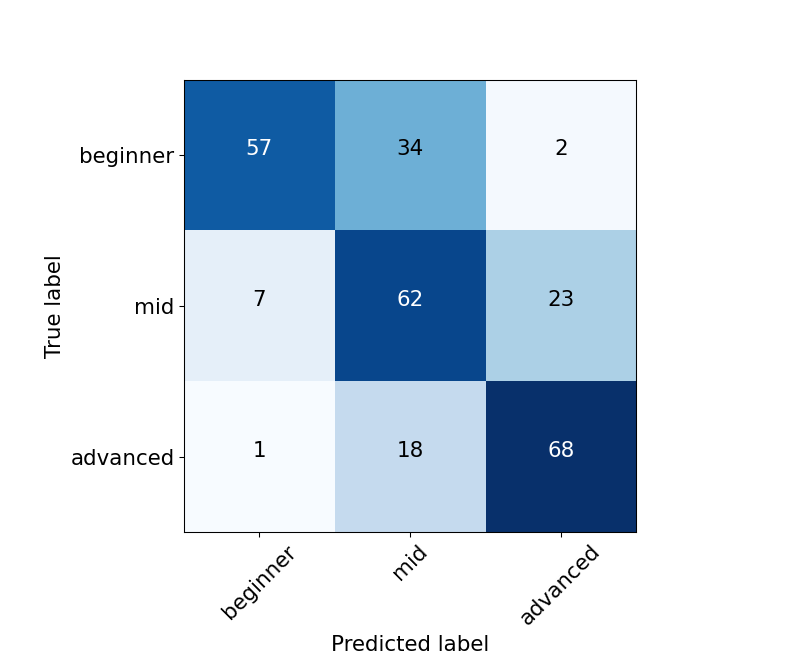}
    \centerline{c) BiLSTM}
\endminipage
 \caption{Confusion matrices for BERT, HAN, and BiLSTM on the OneStopEnglish corpus.} 
   \label{fig:cm_onestopenglish}
\end{figure}

\begin{figure}[t]
\minipage{0.33\textwidth}
  \includegraphics[width=1.0 \linewidth]{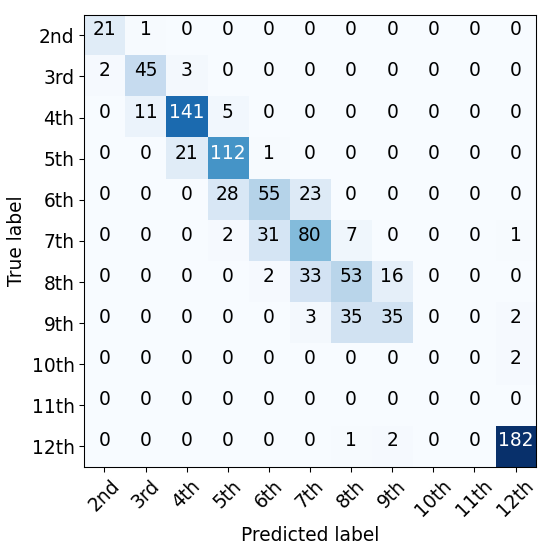}
  \centerline{a) BERT}
\endminipage\hfill
\minipage{0.33\textwidth}
  \includegraphics[width=1.0 \linewidth]{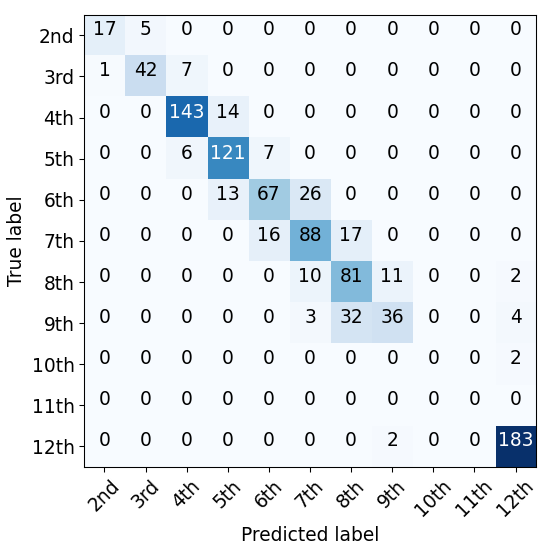}
    \centerline{b) HAN}
\endminipage\hfill
\minipage{0.33\textwidth}
  \includegraphics[width=1.0 \linewidth]{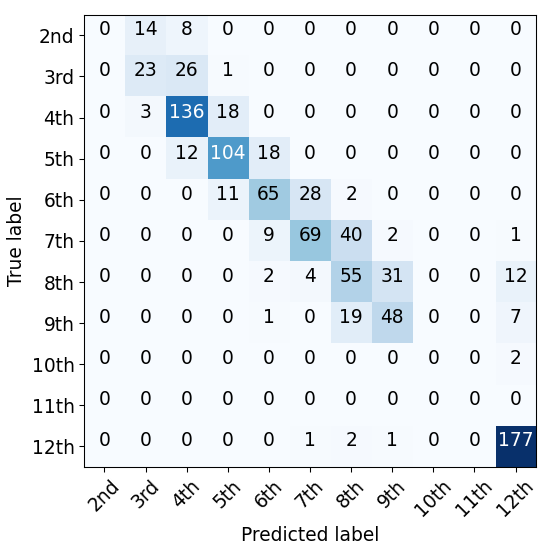}
    \centerline{c) BiLSTM}
\endminipage
 \caption{Confusion matrices for BERT, HAN, and BiLSTM on the Newsela corpus.} 
  \label{fig:cm_newsela}
\end{figure}

\begin{figure}[t]
\minipage{0.33\textwidth}
  \includegraphics[width=1.0 \linewidth]{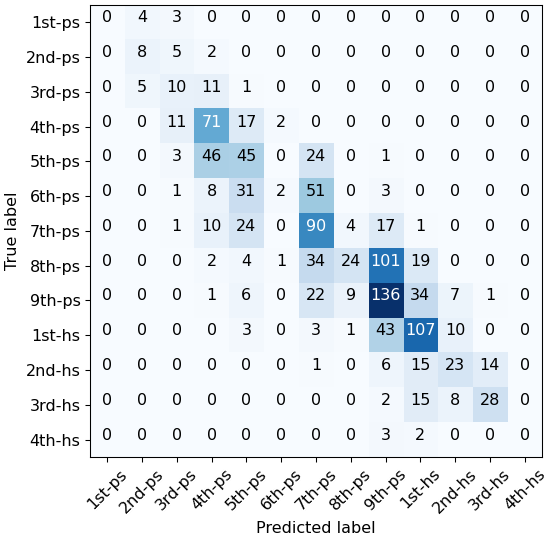}
  \centerline{a) BERT}
\endminipage\hfill
\minipage{0.33\textwidth}
  \includegraphics[width=1.0 \linewidth]{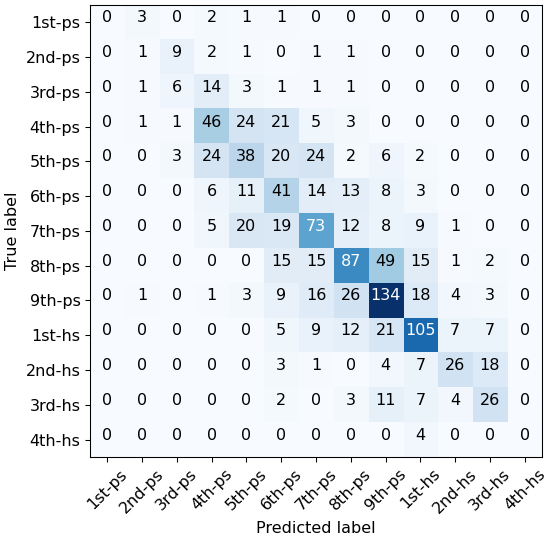}
    \centerline{b) HAN}
\endminipage\hfill
\minipage{0.33\textwidth}
  \includegraphics[width=1.0 \linewidth]{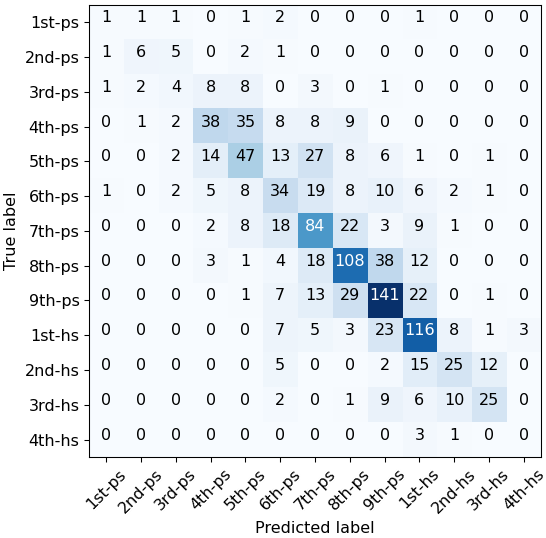}
    \centerline{c) BiLSTM}
\endminipage
 \caption{Confusion matrices for BERT, HAN, and BiLSTM on the Slovenian school books corpus.} 
   \label{fig:cm_ucbeniki}
\end{figure}

On the OneStopEnglish corpus (Figure \ref{fig:cm_onestopenglish}), the BERT classifier, which is performing the worst on this corpus according to all criteria but precision, had the most problems correctly classifying documents from the advanced class, misclassifying about half of the documents. HAN had the most problems with distinguishing documents from the advanced and intermediate class, while BiLSTM classifier classified a disproportionate amount of intermediate documents into the beginner class.

Confusion matrices of all classifiers for the Newsela corpus (Figure \ref{fig:cm_newsela}) follow a similar pattern. Unsurprisingly, no classifier predicted any documents to be in two minority classes (10th and 11th grade) with minimal training examples. As the QWK score has already shown, all classifiers classified a large majority of misclassified instances into neighbouring classes, and costlier mistakes are rare. For example, the best performing HAN classifier altogether misclassified only thirteen examples into non-neighbouring classes.

Confusion matrices for the Slovenian SB corpus (Figure \ref{fig:cm_ucbeniki}) are  similar for all classifiers. The biggest spread of misclassified documents is visible for the classes in the middle of the readability range (from the 4th-grade primary school to the 1st-grade high school). The mistakes, which cause BERT to have lower $F_1$-score and accuracy scores than the other two classifiers, are most likely connected to the misclassification of all but two documents belonging to the school books for the 6th class of the primary school. Nevertheless, a large majority of these documents were misclassified into two neighbouring classes, which explain the high QWK score achieved by the classifier. What negatively affected the QWK scores for HAN and BiLSTM is that the frequency of making costlier mistakes of classifying documents several grades above or below the correct grade is slightly higher for them than for BERT. Nevertheless, even though $F_1$-score results are relatively low on this dataset for all classifiers (BiLSTM achieved the best $F_1$-score of 52.19\%), the QWK scores around or above 80\% and the confusion matrices clearly show that a large majority of misclassified examples were put into classes close to the correct one, suggesting that classification approaches to readability prediction can also be reliably used for Slovenian.

Overall, the classification results suggest that neural networks are a viable option for the supervised readability prediction. Some of the proposed neural approaches managed to outperform state-of-the-art machine learning classifiers that leverage feature engineering \citep{xia2016text, vajjala2018onestopenglish, deutsch2020linguistic} on all corpora where comparisons are available. However, the gains are not substantial, and the choice of an appropriate architecture depends on the properties of the specific dataset.

\section{Conclusion}
\label{sec:conclusion}
We presented a set of novel unsupervised and supervised approaches for determining the readability of documents using deep neural networks. We tested them on several manually labelled English and Slovenian corpora. We argue that deep neural networks are a viable option both for supervised and unsupervised readability prediction and show that the suitability of a specific architecture for the readability task depends on the dataset specifics. 

We demonstrate that neural language models can be successfully employed in the unsupervised setting, since they, in contrast to n-gram models,  capture high-level textual properties and can successfully leverage rich semantic information obtained from the training dataset. However, the results of this study suggest that unsupervised approaches to readability prediction that only take these properties of text into account cannot compete with the shallow lexical sophistication indicators. This is somewhat in line with the findings of the study by \citet{todirascu-etal-2016-cohesive}, who also acknowledged the supremacy of shallow lexical indicators when compared to higher-level discourse features. Nevertheless, combining the components of both neural and traditional readability indicators into the new RSRS (ranked sentence readability score) measure does improve the correlation with human readability scores. 

We argue that the RSRS measure is adaptable, robust, and transferable across languages. The results of the unsupervised experiments show the influence of the language model training set on the performance of the measure. While the results indicate that an exact match between the genres of the train and test sets is not necessary, the text complexity of a train set (i.e. its readability), should be in the lower or middle part of the readability spectrum of the test set for the optimal performance of the measure. This indicates that out of the two high-level text properties that the RSRS measure employs for determining readability, semantic information and long-distance structural information, the latter seems to have more effect on the performance. This is further confirmed by the results of using the general BERT language model for the readability prediction, which show a negative correlation between the language model perplexity and readability, even though the semantic information the model possesses is extensive due to the large training set. 

The functioning of the proposed RSRS measure can be customized and influenced by choice of the training set. This is the desired property since it enables personalization and localization of the readability measure according to the educational needs, language, and topic. The usability of this feature might be limited for under-resourced languages since sufficient amount of documents needed to train a language model that can be used for the task of readability prediction in a specific customized setting might not be available. On the other hand, our experiments on the Slovenian language show, that a relatively small 2.4 million word training corpus for language models is sufficient to outperform traditional readability measures.

The results of the unsupervised approach to readability prediction on the corpus of Slovenian school books are not entirely consistent with the results reported by the previous Slovenian readability study \citep{vskvorcevaluation}, where the authors reported that simple indicators of readability, such as average sentence length, performed quite well. Our results show that the average sentence length performs very competitively on English but ranks badly on Slovenian. This inconsistency in results might be explained by the difference in corpora used for the evaluation of our approaches. While \citet{vskvorcevaluation} conducted experiments on a corpus of magazines for different age groups (which we used for language model training), our experiments were conducted on a corpus of school books, which contains school books for sixteen distinct school subjects with very different topics ranging from literature, music and history to math, biology and chemistry. As was already shown in \citet{sheehan2013two}, the variance in genres and covered topics has an important effect on the ranking and performance of different readability measures. Further experiments on other Slovenian datasets are required to confirm this hypothesis.  

In the supervised approach to determining readability, we show that the proposed neural classifiers can either outperform or at least compare with state-of-the-art approaches leveraging extensive feature engineering as well as previously employed neural models on all corpora where comparison data is available. While the improved performance and elimination of work required for manual feature engineering are desirable, on the downside, neural approaches tend to decrease the interpretability and explainability of the readability prediction. Interpretability and explainability are especially important for educational applications \citep{madnani2018automated,sheehan2014textevaluator}, where the users of such technology (educators, teachers, researchers, etc.) often need to understand what causes one text to be judged as more readable than the other and according to which dimensions. Therefore in the future, we will explore the possibilities of explaining the readability predictions of the proposed neural classifier with the help of general explanation techniques such as SHAP \citep{Lundberg2017}, or the attention mechanism \citep{vaswani2017attention}, which can be analyzed and visualized and can offer valuable insights into inner workings of the system.

Another issue worth discussing is the trade-off between performance gains we can achieve by employing computationally demanding neural networks on the one side and the elimination of work on the other. For example, on the OneStopEnglish corpus, we report the accuracy of 78.72\% when HAN is employed, while \citet{vajjala2018onestopenglish} report an accuracy of 78.13\% with their classifier employing 155 hand-crafted features. While it might be worth opting for a neural network in order to avoid extensive manual feature engineering, on the other hand, the same study by \citet{vajjala2018onestopenglish} also reports that just by employing generic text classification features, 2-5 character n-grams, they obtained the accuracy of 77.25\%. Considering this, one might argue that, depending on the use case, it might not be worth dedicating significantly more time, work or computational resources for an improvement of slightly more than 1\%, especially if this also decreases the overall interpretability of the prediction.  

The performance of different classifiers varies across different corpora. The major factor proved to be the length of documents in the datasets. The HAN architecture, which tends to be well equipped to handle long-distance hierarchical text structures, performs the best on these datasets. On the other hand, in terms of QWK measure, BERT offers significantly better performance on datasets that contain shorter documents, such as WeeBit and Slovenian SB. As was already explained in Section \ref{sec-results-supervised}, a large majority of OneStopEnglish and Newsela documents need to be truncated in order to satisfy the BERT's limitation of 512 byte-pair tokens. While it is reasonable to assume that the truncation and the consequential loss of information do have a detrimental effect on the performance of the classifier, the extent of this effect is still unclear. The problem of truncation also raises the question of what is the minimum required length of a text for a reliable assessment of readability and if there exists a length threshold, above which having more text does not influence the performance of a classifier in a significant manner. We plan to assess this in future work thoroughly. Another related line of research we plan to pursue in the future is the employment of novel algorithms, such as Longformer \citep{beltagy2020longformer} and Linformer \citep{wang2020linformer}, in which the attention mechanism scales linearly with the sequence length, making it feasible to process documents of thousands of tokens. We will check if applying these two algorithms on the readability datasets with longer documents can further improve the state-of-the-art.

The other main difference between WeeBit and Slovenian SB datasets on the one hand, and Newsela and OneStopEnglish datasets on the other, is that they are not parallel corpora, which means there can be substantial semantic differences between the readability classes in these two corpora. It seems that pretraining BERT as a language model allows for better exploitation of these differences, which leads to better performance. However, this reliance on semantic information might badly affect the performance of transfer learning based models on parallel corpora, since the semantic differences between classes in these corpora are much more subtle. We plan to assess the influence of available semantic information on the performance of different classification models in the future.

The differences in performance between classifiers on different corpora suggest that tested classifiers take different types of information into account. Provided that this hypothesis is correct, some gains in performance might be achieved if the classifiers are combined. We plan to test a neural ensemble approach for the task of predicting readability in the future.

While this study mostly focused on multi-lingual and multi-genre readability prediction, in the future, we also plan to test the cross-corpus, cross-genre and cross-language transferability of the proposed supervised and unsupervised approaches. This requires new readability datasets for different languages and genres which are currently rare or not publicly available. On the other hand, this type of research will be capable of further determining the role of genre in the readability prediction and might open an opportunity to improve the proposed unsupervised readability score further.  

\begin{acknowledgments}
The research was financially supported by the European social fund and Republic of Slovenia, Ministry of Education, Science and Sport through project Quality of Slovene textbooks (KaU\v{C}). The work was also supported by the Slovenian Research Agency (ARRS) through core research programmes P6-0411 and P2-0103, and the projects Terminology and knowledge frames across languages (J6-9372) and Quantitative and qualitative analysis of the unregulated corporate financial reporting (J5-2554). 
This work has also received funding from the European Union's Horizon 2020 research and innovation programme under grant agreement No 825153 (EMBEDDIA). The results of this publication reflect only the authors' views, and the EC is not responsible for any use that may be made of the information it contains.
\end{acknowledgments}

\starttwocolumn
\bibliography{main,EmbeddiaRefs}

\end{document}